\documentclass[runningheads]{llncs}
\usepackage{graphicx}
\usepackage{comment}
\usepackage{amsmath,amssymb} 
\usepackage{color}

\usepackage{mathtools}
\usepackage{epsfig}

\usepackage{subcaption}
\usepackage{multirow}
\usepackage{tabularx}
\usepackage{booktabs}
\usepackage[export]{adjustbox}

\usepackage{cite}
\usepackage{url}

\usepackage{array}
\newcolumntype{P}[1]{>{\centering\arraybackslash}p{#1}}

\newcommand{\etal}{\textit{et al}. }

\begin{document}
\pagestyle{headings}
\mainmatter
\def\ECCVSubNumber{6719}  

\title{Jointly De-biasing Face Recognition and Demographic Attribute Estimation} 

\titlerunning{DebFace}
%
%
\author{Sixue Gong\quad\quad Xiaoming Liu\quad\quad Anil K. Jain\\
{\tt\small \{gongsixu, liuxm, jain\}@msu.edu}
}
\authorrunning{Sixue Gong et al.}
%
\institute{Michigan State University
}
\maketitle

\begin{abstract}
We address the problem of bias in automated face recognition and demographic attribute estimation algorithms, where errors are lower on certain cohorts belonging to specific demographic groups. We present a novel de-biasing adversarial network (DebFace) that learns to extract disentangled feature representations for both unbiased face recognition and demographics estimation. The proposed network consists of one identity classifier and three demographic classifiers (for gender, age, and race) that are trained to distinguish identity and demographic attributes, respectively. Adversarial learning is adopted to minimize correlation among feature factors so as to abate bias influence from other factors. We also design a new scheme to combine demographics with identity features to strengthen robustness of face representation in different demographic groups. The experimental results show that our approach is able to reduce bias in face recognition as well as demographics estimation while achieving state-of-the-art performance.

\end{abstract}

\section{Introduction}

Automated face recognition has achieved remarkable success with the rapid developments of deep learning algorithms. 
Despite the improvement in the accuracy of face recognition, one topic is of significance. Does a face recognition system perform equally well in different demographic groups? In fact, it has been observed that many face recognition systems have lower performance in certain demographic groups than others~\cite{howard2019effect, klare2012face, grother2019frvt}. Such face recognition systems are said to be {\it biased} in terms of demographics. 

In a time when face recognition systems are being deployed in the real world for societal benefit, this type of bias~\footnote{This is different from the notion of machine learning bias, defined as ``any basis for choosing one generalization [hypothesis] over another, other than strict consistency with the observed training instances"~\cite{dietterich1995machine}.} is not acceptable. Why does the bias problem exist in face recognition systems? First, state-of-the-art (SOTA) face recognition methods are based on deep learning which requires a large collection of face images for training. Inevitably the distribution of training data has a great impact on the performance of the resultant deep learning models. It is well understood that face datasets exhibit imbalanced demographic distributions where the number of faces in each cohort is unequal. Previous studies have shown that models trained with imbalanced datasets lead to biased discrimination~\cite{bolukbasi2016man, torralba2011unbiased}. Secondly, the goal of deep face recognition is to map the input face image to a target feature vector with high discriminative power. The bias in the mapping function will result in feature vectors with lower discriminability for certain demographic groups. Moreover, Klare~\etal~\cite{klare2012face} show the errors that are inherent to some demographics by studying non-trainable face recognition algorithms.

To address the bias issue, data re-sampling methods have been exploited to balance the data distribution by under-sampling the majority~\cite{drummond2003c4} or over-sampling the minority classes~\cite{chawla2002smote, mullick2019generative}. Despite its simplicity, valuable information may be removed by under-sampling, and over-sampling may introduce noisy samples. Naively training on a balanced dataset can still lead to bias~\cite{wang2020mitigating}. Another common option for imbalanced data training is cost-sensitive learning that assigns weights to different classes based on (i) their frequency or (ii) the effective number of samples~\cite{cao2019learning, cui2019class}. To eschew the overfitting of Deep Neural Network (DNN) to minority classes, hinge loss is often used to increase margins among classification decision boundaries~\cite{hayat2019max, khan2019striking}. The aforementioned methods have also been adopted for face recognition and attribute prediction on imbalanced datasets~\cite{huang2019deep, wang2019deep}. 
However, such face recognition studies only concern bias in terms of {\it identity}, rather than our focus of {\it demographic bias}. 

\begin{figure}[t!]
	    \captionsetup{font=footnotesize}
	    \centering
	    \begin{subfigure}[b]{0.33\linewidth}
	    \includegraphics[width=0.95\linewidth]{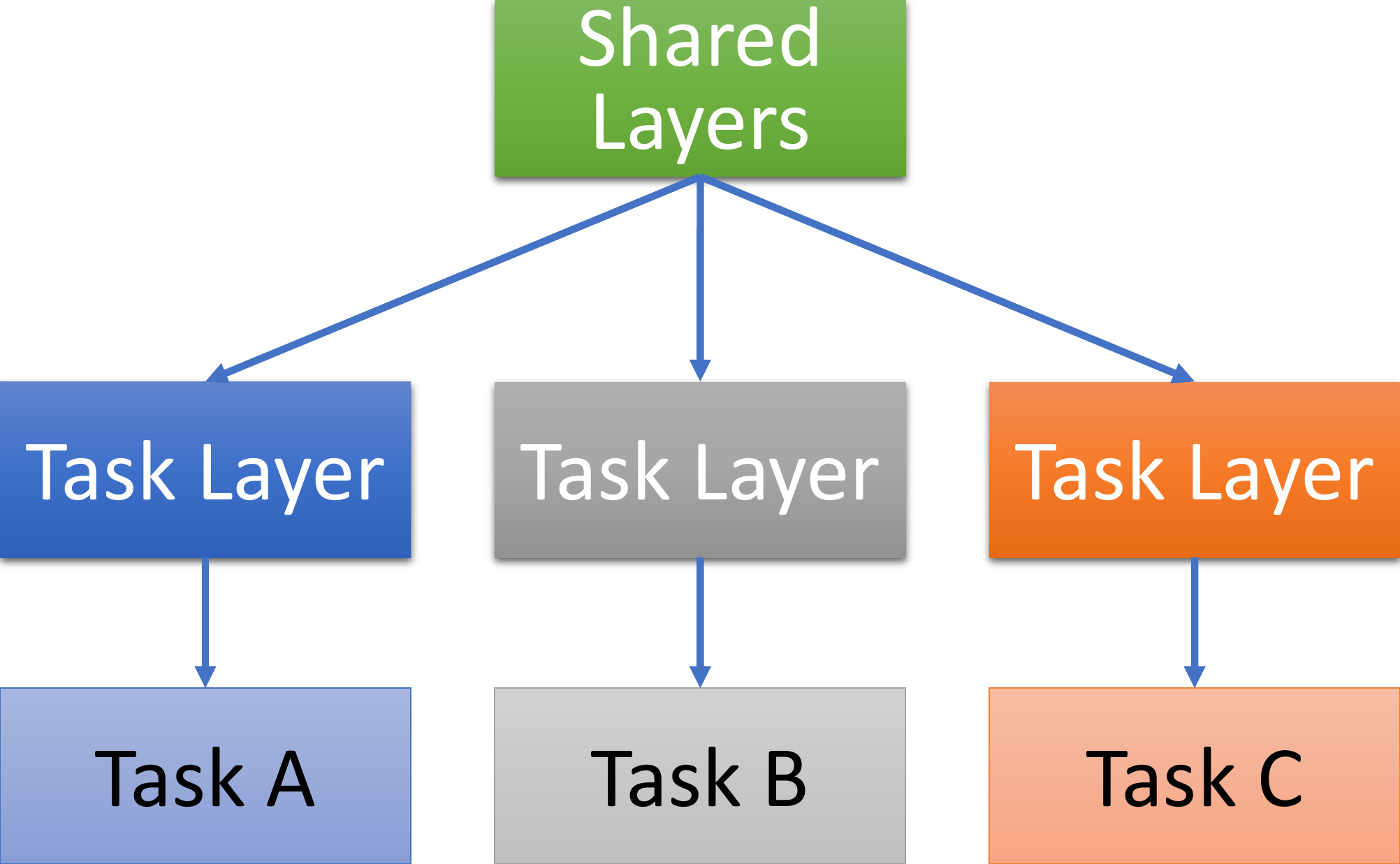}
	    \caption{{\scriptsize Multi-task learning}}
	    \label{fig:multi-task}
	    \end{subfigure}\hfill		
		\begin{subfigure}[b]{0.33\linewidth}
		\includegraphics[width=0.95\linewidth]{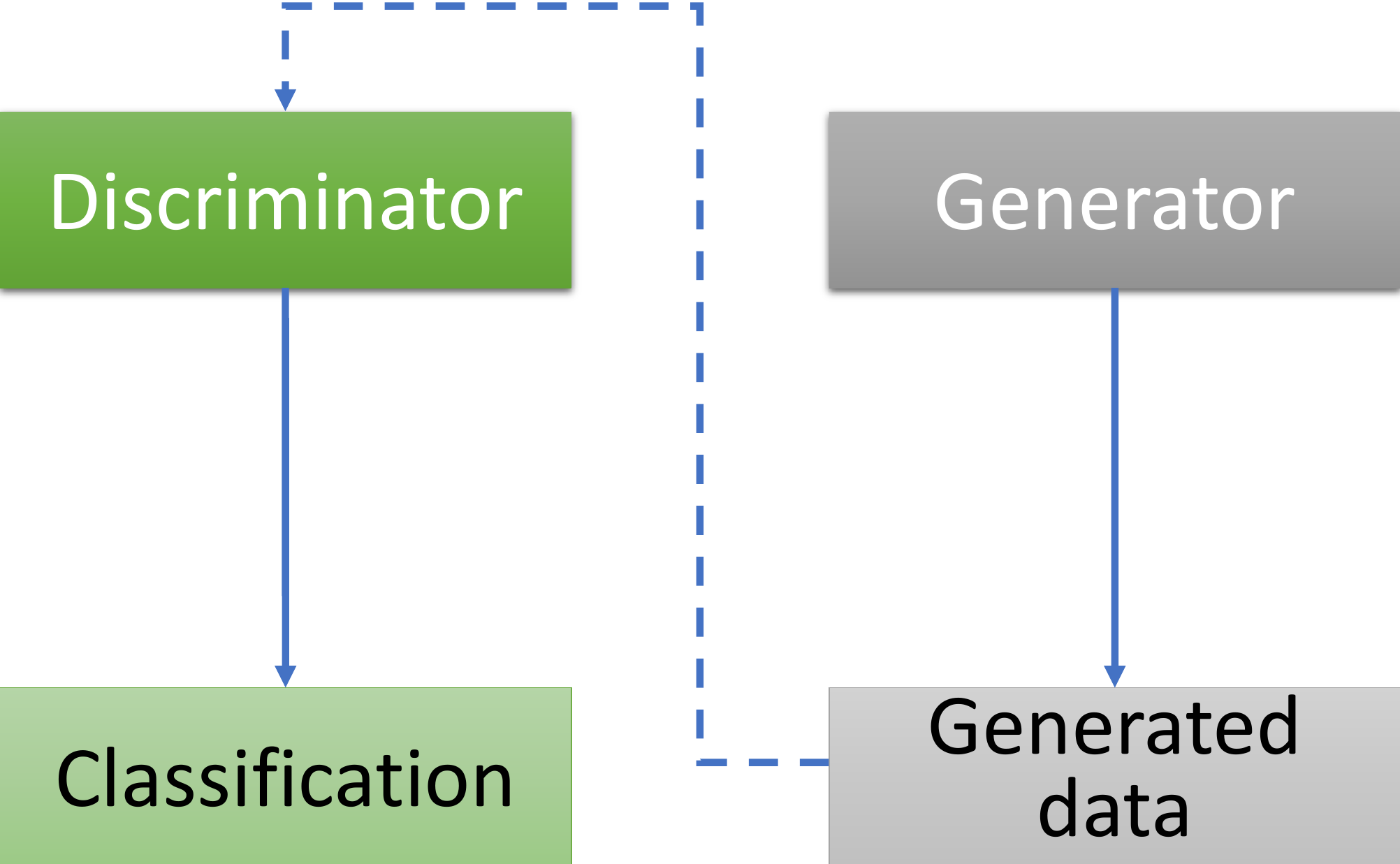}
	    \caption{{\scriptsize Adversarial learning}}
	    \label{fig:gan}
	    \end{subfigure}\hfill    
	    \begin{subfigure}[b]{0.34\linewidth}
	    \includegraphics[width=0.95\linewidth]{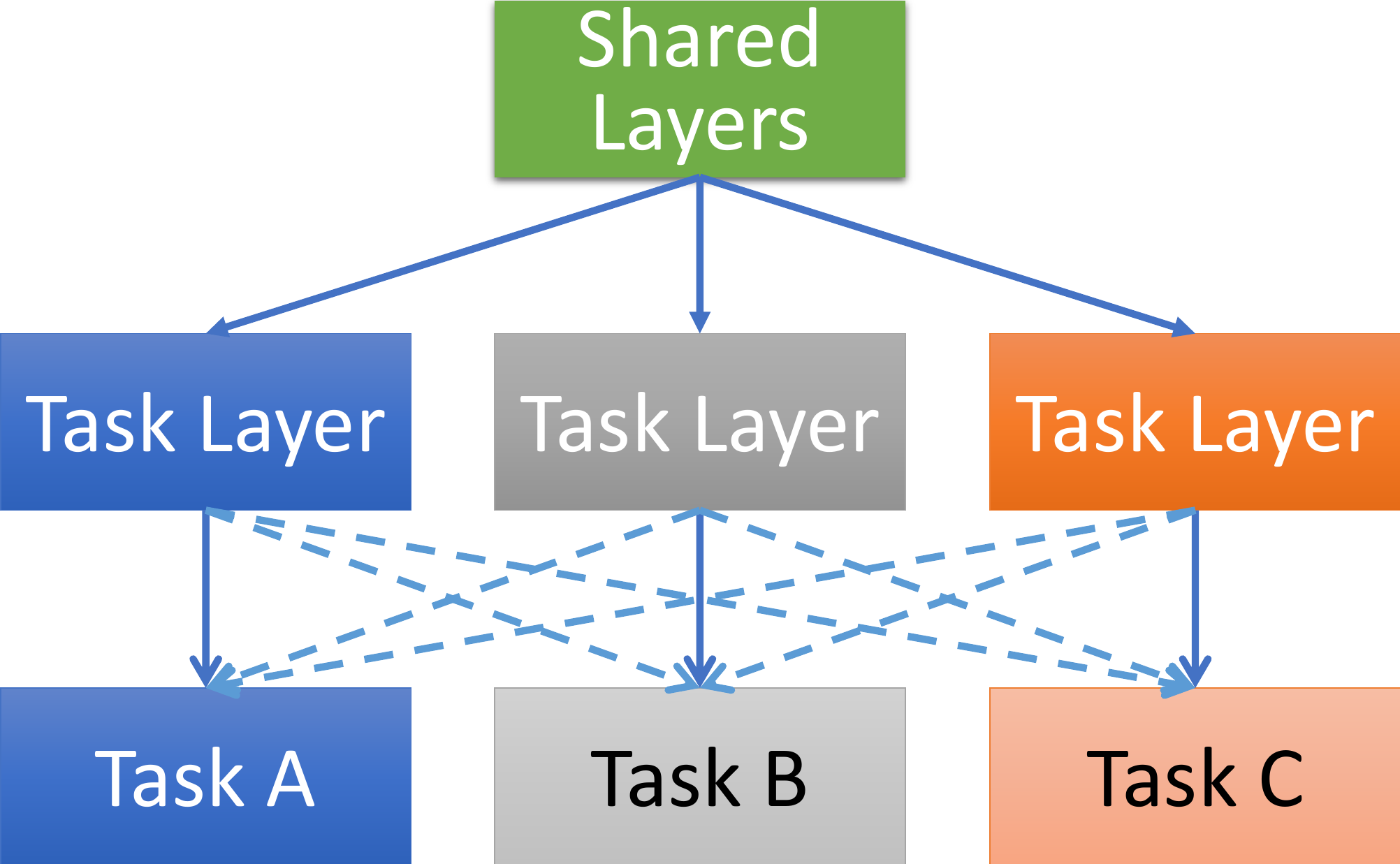}
	    \caption{{\scriptsize DebFace}}
	    \label{fig:adv_train}
	    \end{subfigure}\\
	    
	    \caption{\footnotesize{Methods to learn different tasks simultaneously. Solid lines are typical feature flow in CNN, while dash lines are adversarial losses.}}
	    
	    \label{fig:related_work}
\end{figure}

In this paper, we propose a framework to address the influence of bias on face recognition and demographic attribute estimation. 
In typical deep learning based face recognition frameworks, the large capacity of DNN enables the face representations to embed demographic details, including gender, race, and age~\cite{abadi2016deep, fredrikson2015model}. 
Thus, the biased demographic information is transmitted from the training dataset to the output representations. 
To tackle this issue, we assume that if the face representation does not carry discriminative information of demographic attributes, it would be unbiased in terms of demographics. 
Given this assumption, one common way to remove demographic information from face representations is to perform feature disentanglement via adversarial learning  (Fig.~\ref{fig:gan}).
That is, the classifier of demographic attributes can be used to encourage the identity representation to {\it not} carry demographic information.
However, one issue of this common approach is that, the demographic classifier itself could be biased ({\it e.g.}, the race classifier could be biased on gender), and hence it will act differently while disentangling faces of different cohorts.
This is clearly undesirable as it leads to demographic biased identity representation.

To resolve the chicken-and-egg problem, we propose to {\it jointly} learn unbiased representations for both the identity and demographic attributes.
Specifically, starting from a multi-task learning framework that learns disentangled feature representations of gender, age, race, and identity, respectively, we request the classifier of each task to act as adversarial supervision for the other tasks ({\it e.g.}, the dash arrows in Fig.~\ref{fig:adv_train}).
These four classifiers help each other to achieve better feature disentanglement, resulting in unbiased feature representations for both the identity and demographic attributes.
As shown in Fig.~\ref{fig:related_work}, our framework is in sharp contrast to either multi-task learning or adversarial learning.


Moreover, since the features are disentangled into the demographic and identity, our face representations also contribute to privacy-preserving applications. 
It is worth noticing that such identity representations contain little demographic information, which could undermine the recognition competence since demographic features are {\it part} of identity-related facial appearance. 
To retain the recognition accuracy on demographic biased face datasets, we propose another network that combines the demographic features with the demographic-free identity features to generate a new identity representation for face recognition. 

The key contributions and findings of the paper are: 

$\diamond$   A thorough analysis of deep learning based face recognition performance on three different demographics: (i) gender, (ii) age, and (iii) race. 

$\diamond$   A de-biasing face recognition framework, called DebFace, that generates disentangled representations for both identity and demographics recognition while jointly removing discriminative information from other counterparts. 

$\diamond$   The identity representation from DebFace (DebFace-ID) shows lower bias on different demographic cohorts and also achieves SOTA face verification results on demographic-unbiased face recognition. 

$\diamond$  The demographic attribute estimations via DebFace are less biased across other demographic cohorts. 

$\diamond$   Combining ID with demographics results in more discriminative features for face recognition on biased datasets. 

\section{Related Work}
\paragraph{Face Recognition on Imbalanced Training Data}
Previous efforts on face recognition aim to tackle class imbalance problem on training data. 
For example, in prior-DNN era, Zhang~\etal~\cite{zhang2009cost} propose a cost-sensitive learning framework to reduce misclassification rate of face identification. 
To correct the skew of separating hyperplanes of SVM on imbalanced data, Liu~\etal~\cite{liu2007face} propose Margin-Based Adaptive Fuzzy SVM that obtains a lower generalization error bound. 
In the DNN era, face recognition models are trained on large-scale face datasets with highly-imbalanced class distribution~\cite{yin2019feature, zhang2017range}. 
Range Loss~\cite{zhang2017range} learns a robust face representation that makes the most use of every training sample. 
To mitigate the impact of insufficient class samples, center-based feature transfer learning~\cite{yin2019feature} and large margin feature augmentation~\cite{wang2019deep} are proposed to augment features of minority identities and equalize class distribution. 
Despite their effectiveness, these studies ignore the influence of demographic imbalance on the dataset, which may lead to demographic bias. 
For instance, The FRVT 2019~\cite{grother2019frvt} shows the demographic bias of over 100 face recognition algorithms.
To uncover deep learning bias, Alexander~\etal~\cite{amini2019uncovering} develop an algorithm to mitigate the hidden biases within training data.
Wang~\etal~\cite{Wang_2019_ICCV} propose a domain adaptation network to reduce racial bias in face recognition. They recently extended their work using reinforcement learning to find optimal margins of additive angular margin based loss functions for different races~\cite{wang2020mitigating}.
To our knowledge, no studies have tackled the challenge of de-biasing demographic bias in DNN-based face recognition and demographic attribute estimation algorithms.

\paragraph{Adversarial Learning and Disentangled Representation}
Adversarial learning~\cite{schmidhuber1992learning} has been well explored in many computer vision applications. 
For example, Generative Adversarial Networks (GANs)~\cite{goodfellow2014generative} employ adversarial learning to train a generator by competing with a discriminator that distinguishes real images from synthetic ones. Adversarial learning has also been applied to domain adaptation~\cite{tzeng2015simultaneous, tzeng2017adversarial, long2018conditional, tao2019minimax}. 
A problem of current interest is to learn interpretable representations with semantic meaning~\cite{towards-interpretable-face-recognition}. 
Many studies have been learning factors of variations in the data by supervised learning~\cite{liu2018multi,liu2018exploring, disentangled-representation-learning-gan-for-pose-invariant-face-recognition,representation-learning-by-rotating-your-faces,disentangling-features-in-3d-face-shapes-for-joint-face-reconstruction-and-recognition}, or semi-supervised/unsupervised learning~\cite{kim2018disentangling, narayanaswamy2017learning, locatello2018challenging,gait-recognition-via-disentangled-representation-learning}, referred to as disentangled representation. 
For supervised disentangled feature learning, adversarial networks are utilized to extract features that only contain discriminative information of a target task. 
For face recognition, Liu~\etal~\cite{liu2018exploring} propose a disentangled representation by training an adversarial autoencoder to extract features that can capture identity discrimination and its complementary knowledge. 
In contrast, our proposed DebFace differs from prior works in that each branch of a multi-task network acts as both a generator and discriminators of other branches (Fig.~\ref{fig:adv_train}).

\section{Methodology}
\subsection{Problem Definition}
The ultimate goal of unbiased face recognition is that, given a face recognition system, no statistically significant difference among the performance in different categories of face images.
Despite the research on pose-invariant face recognition that aims for equal performance on all poses~\cite{towards-large-pose-face-frontalization-in-the-wild,representation-learning-by-rotating-your-faces}, we believe that it is inappropriate to define variations like pose, illumination, or resolution, as the categories.
These are instantaneous {\it image-related} variations with intrinsic bias. 
E.g., large-pose or low-resolution faces are inherently harder to be recognized than frontal-view high-resolution faces.

Rather, we would like to define {\it subject-related} properties such as demographic attributes as the categories.
\textit{A face recognition system is \textbf{biased} if it performs worse on certain demographic cohorts.} 
For practical applications, it is important to consider what demographic biases may exist, and whether these are intrinsic biases across demographic cohorts or algorithmic biases derived from the algorithm itself. 
This motivates us to analyze the demographic influence on face recognition performance and strive to reduce algorithmic bias for face recognition systems.
One may achieve this by training on a dataset containing uniform samples over the cohort space. 
However, the demographic distribution of a dataset is often imbalanced and underrepresents demographic minorities while overrepresenting majorities. 
Naively re-sampling a balanced training dataset may still induce bias since the diversity of latent variables is different across cohorts and the instances cannot be treated fairly during training. 
To mitigate demographic bias, we propose a face de-biasing framework that jointly reduces mutual bias over all demographics and identities while disentangling face representations into gender, age, race, and demographic-free identity in the mean time.

\subsection{Algorithm Design}
The proposed network takes advantage of the relationship between demographics and face identities. 
On one hand, demographic characteristics are highly correlated to face features. 
On the other hand, demographic attributes are heterogeneous in terms of data type and semantics~\cite{Han2017Heterogeneous}. 
A male person, for example, is not necessarily of a certain age or of a certain race.
Accordingly, we present a framework that jointly generates demographic features and identity features from a single face image by considering both the aforementioned attribute correlation and attribute heterogeneity in a DNN.

While our main goal is to mitigate demographic bias from face representation, we observe that demographic estimations are biased as well (see Fig.~\ref{fig:demog_bias}). 
How can we remove the bias of face recognition when demographic estimations themselves are biased? Cook~\etal~\cite{cook2019demographic} investigated this effect and found the performance of face recognition is affected by multiple demographic covariates.
We propose a de-biasing network, DebFace, that disentangles the representation into gender (DebFace-G), age (DebFace-A), race (DebFace-R), and identity (DebFace-ID), to decrease bias of both face recognition and demographic estimations.  
Using adversarial learning, the proposed method is capable of jointly learning multiple discriminative representations while ensuring that each classifier cannot distinguish among classes through non-corresponding representations. 

Though less biased, DebFace-ID loses demographic cues that are useful for identification. 
In particular, race and gender are two critical components that constitute face patterns. 
Hence, we desire to incorporate race and gender with DebFace-ID to obtain a more integrated face representation. 
We employ a light-weight fully-connected network to aggregate the representations into a face representation (DemoID) with the same dimensionality as DebFace-ID.

\begin{figure}[t!]
    \centering
    \includegraphics[width=\linewidth]{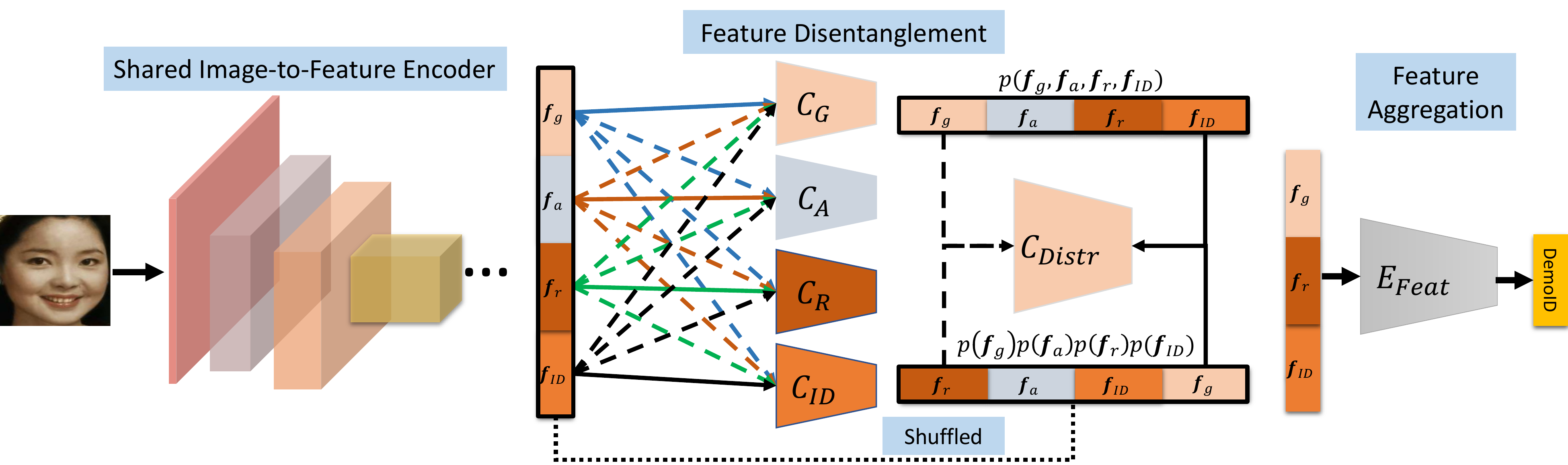}
    \caption{\footnotesize{Overview of the proposed De-biasing face (DebFace) network. DebFace is composed of three major blocks, {\it i.e.}, a shared feature encoding block, a feature disentangling block, and a feature aggregation block. The solid arrows represent the forward inference, and the dashed arrows stand for adversarial training. During inference, either DebFace-ID ({\it i.e.}, $\mathbf{f}_{ID}$) or DemoID can be used for face matching given the desired trade-off between biasness and accuracy. }}
    \label{fig:framework}
    \textbf{}
\end{figure}

\subsection{Network Architecture}
\label{sec:net}
Figure~\ref{fig:framework} gives an overview of the proposed DebFace network. It consists of four components: the shared image-to-feature encoder $E_{Img}$, the four attribute classifiers (including gender $C_{G}$, age $C_A$, race $C_R$, and identity $C_{ID}$), the distribution classifier $C_{Distr}$, and the feature aggregation network $E_{Feat}$.
We assume access to $N$ labeled training samples $\{ (\mathbf{x}^{(i)},y_g^{(i)}, y_a^{(i)}, y_r^{(i)}, y_{id}^{(i)} ) \}_{i=1}^N$. Our approach takes an image $\mathbf{x}^{(i)}$ as the input of $E_{Img}$. The encoder projects $\mathbf{x}^{(i)}$ to its feature representation $E_{Img}(\mathbf{x}^{(i)})$. The feature representation is then decoupled into four $D$-dimensional feature vectors, gender $\mathbf{f}_g^{(i)}$, age $\mathbf{f}_a^{(i)}$, race $\mathbf{f}_r^{(i)}$, and identity $\mathbf{f}_{ID}^{(i)}$, respectively. Next, each attribute classifier operates the corresponding feature vector to correctly classify the target attribute by optimizing  parameters of both $E_{Img}$ and the respective classifier $C_{*}$. \sloppy For a demographic attribute with $K$ categories, the learning objective $\mathcal{L}_{C_{Demo}}(\mathbf{x},y_{Demo};E_{Img}, C_{Demo})$ is the standard cross entropy loss function.
For the $n-$identity classification, we adopt AM-Softmax~\cite{wang2018additive} as the objective function $\mathcal{L}_{C_{ID}}(\mathbf{x},y_{id};E_{Img},C_{ID})$. 
\sloppy To de-bias all of the feature representations, adversarial loss $\mathcal{L}_{Adv}(\mathbf{x},y_{Demo},y_{id};E_{Img},C_{Demo},C_{ID})$ is applied to the above four classifiers such that each of them will not be able to predict correct labels when operating irrelevant feature vectors. Specifically, given a classifier, the remaining three attribute feature vectors are imposed on it and attempt to mislead the classifier by only optimizing the parameters of $E_{Img}$. 
To further improve the disentanglement, we also reduce the mutual information among the attribute features by introducing a distribution classifier $C_{Distr}$. 
$C_{Distr}$ is trained to identify whether an input representation is sampled from the joint distribution $p(\mathbf{f}_g, \mathbf{f}_a, \mathbf{f}_r, \mathbf{f}_{ID})$ or the multiplication of margin distributions $p(\mathbf{f}_g)p(\mathbf{f}_a)p(\mathbf{f}_r)p(\mathbf{f}_{ID})$ via a binary cross entropy loss $\mathcal{L}_{C_{Distr}}(\mathbf{x},y_{Distr};E_{Img},C_{Distr})$, where $y_{Distr}$ is the distribution label. 
Similar to adversarial loss, a factorization objective function $\mathcal{L}_{Fact}(\mathbf{x},y_{Distr};E_{Img},C_{Distr})$ is utilized to restrain the $C_{Distr}$ from distinguishing the real distribution and thus minimizes the mutual information of the four attribute representations. Both adversarial loss and factorization loss are detailed in Sec.~\ref{sec:adv}.
Altogether,  DebFace endeavors to minimize the joint loss:
\begin{equation}
\label{eq:lossFunc}
\begin{split}
\mathcal{L}(\mathbf{x}, y_{Demo}, & y_{id},y_{Distr};E_{Img},C_{Demo},C_{ID},C_{Distr}) = \\
& \mathcal{L}_{C_{Demo}}(\mathbf{x},y_{Demo};E_{Img}, C_{Demo}) \\
& + \mathcal{L}_{C_{ID}}(\mathbf{x},y_{id};E_{Img},C_{ID}) \\
& + \mathcal{L}_{C_{Distr}}(\mathbf{x},y_{Distr};E_{Img},C_{Distr}) \\
& + \lambda \mathcal{L}_{Adv}(\mathbf{x},y_{Demo},y_{id};E_{Img},C_{Demo},C_{ID}) \\
& + \nu \mathcal{L}_{Fact}(\mathbf{x},y_{Distr};E_{Img},C_{Distr}),
\end{split}
\end{equation}
where $\lambda$ and $\nu$ are hyper-parameters determining how much the representation is decomposed and decorrelated in each training iteration.

The discriminative demographic features in DebFace-ID are weakened by removing demographic information.
Fortunately, our de-biasing network preserves all pertinent demographic features in a disentangled way. 
Basically, we train another multilayer perceptron (MLP) $E_{Feat}$ to aggregate DebFace-ID and the demographic embeddings into a unified face representation DemoID. 
Since age generally does not pertain to a person's identity, we only consider gender and race as the identity-informative attributes. 
The aggregated embedding, 
 $\mathbf{f}_{DemoID} = E_{feat}(\mathbf{f}_{ID}, \mathbf{f}_g, \mathbf{f}_r)$, is supervised by an identity-based triplet loss:
 
\noindent \resizebox{0.99\linewidth}{!}
{
\begin{minipage}{\linewidth}
\begin{eqnarray}
\mathcal{L}_{E_{Feat}} = \frac{1}{M} \sum_{i=1}^M [{\| \mathbf{f}_{DemoID^{a}}^{(i)} - \mathbf{f}_{DemoID^{p}}^{(i)} \|}^2_2 - {\| \mathbf{f}_{DemoID^{a}}^{(i)} - \mathbf{f}_{DemoID^{n}}^{(i)} \|}^2_2 + \alpha]_+,  \hspace{5mm}
\label{eqn:L_adv}
\end{eqnarray}
\end{minipage}
}

\noindent where  $\{ \mathbf{f}_{DemoID^{a}}^{(i)}, \mathbf{f}_{DemoID^{p}}^{(i)}, \mathbf{f}_{DemoID^{n}}^{(i)} \}$ is the $i^{th}$ triplet consisting of an anchor, a positive, and a negative DemoID representation, $M$ is the number of hard triplets in a mini-batch. $[x]_+ = \max (0, x)$, and $\alpha$ is the margin.

\subsection{Adversarial Training and Disentanglement}
\label{sec:adv}
As discussed in Sec.~\ref{sec:net}, the adversarial loss aims to minimize the task-independent information semantically, while the factorization loss strives to dwindle the interfering information statistically. 
We employ both losses to disentangle the representation extracted by $E_{Img}$. 
We introduce the adversarial loss as a means to learn a representation that is invariant in terms of certain attributes, where a classifier trained on it cannot correctly classify those attributes using that representation. 
We take one of the attributes, {\it e.g.}, gender, as an example to illustrate the adversarial objective. 
First of all, for a demographic representation $\mathbf{f}_{Demo}$, we learn a gender classifier on $\mathbf{f}_{Demo}$ by optimizing the classification loss $\mathcal{L}_{C_{G}}(\mathbf{x},y_{Demo};E_{Img}, C_{G})$. 
Secondly, for the same gender classifier, we intend to maximize the chaos of the predicted distribution~\cite{attribute-preserved-face-de-identification}. 
It is well known that a uniform distribution has the highest entropy and presents the most randomness. 
Hence, we train the classifier to predict the probability distribution as close as possible to a uniform distribution over the category space by minimizing the cross entropy:

\noindent \resizebox{0.86\linewidth}{!}{
\begin{minipage}{\linewidth}
\begin{eqnarray}
\mathcal{L}_{Adv}^{G} (\mathbf{x},y_{Demo},y_{id};E_{Img},C_{G}) = -\sum_{k=1}^{K_G} \frac{1}{K_G} \cdot (\log \frac{e^{C_{G}(\mathbf{f}_{Demo})_{k}}}{\sum_{j=1}^{K_G} e^{C_{G}(\mathbf{f}_{Demo})_{j}}} + \log \frac{e^{C_{G}(\mathbf{f}_{ID})_{k}}}{\sum_{j=1}^{K_G} e^{C_{G}(\mathbf{f}_{ID})_{j}}}),  \hspace{5mm}
\label{eqn:L_adv}
\end{eqnarray}
\end{minipage}
}

\noindent where $K_G$ is the number of categories in gender~\footnote{In our case, $K_G=2$, {\it i.e.}, male and female.}, and the ground-truth label is no longer an one-hot vector, but a $K_G$-dimensional vector with all elements being  $\frac{1}{K_G}$. 
The above loss function corresponds to the dash lines in Fig.~\ref{fig:framework}. It strives for gender-invariance by finding a representation that makes the gender classifier $C_{G}$ perform poorly. We minimize the adversarial loss by only updating parameters in $E_{Img}$.

We further decorrelate the representations by reducing the mutual information across attributes. 
By definition, the mutual information is the relative entropy (KL divergence) between the joint distribution and the product distribution. 
To increase uncorrelation, we add a distribution classifier $C_{Distr}$ that is trained to simply perform a binary classification using $\mathcal{L}_{C_{Distr}}(\mathbf{x},y_{Distr};E_{Img},C_{Distr})$ on samples $\mathbf{f}_{Distr}$ from both the joint distribution and dot product distribution. 
Similar to adversarial learning, we factorize the representations by tricking the classifier via the same samples so that the predictions are close to random guesses,  
 \begin{equation}
 \mathcal{L}_{Fact}(\mathbf{x},y_{Distr}; E_{Img},C_{Distr}) = -\sum_{i=1}^2 \frac{1}{2} \log \frac{e^{C_{Distr}(\mathbf{f}_{Distr})_{i}}}{\sum_{j=1}^{2} e^{C_{Distr}(\mathbf{f}_{Distr})_{j}}}.
 \end{equation}
In each mini-batch, we consider $E_{Img}(\mathbf{x})$ as samples of the joint distribution $p(\mathbf{f}_g, \mathbf{f}_a, \mathbf{f}_r, \mathbf{f}_{ID})$. We randomly shuffle feature vectors of each attribute, and re-concatenate them into $4D$-dimension, which are approximated as samples of the product distribution $p(\mathbf{f}_g)p(\mathbf{f}_a)p(\mathbf{f}_r)p(\mathbf{f}_{ID})$. During factorization, we only update $E_{Img}$ to minimize mutual information between decomposed features.

\section{Experiments}
\subsection{Datasets and Pre-processing}
\label{sec:datasets}
We utilize $15$ total face datasets in this work, for learning the demographic estimation models, the baseline face recognition model, DebFace model as well as their evaluation.  To be specific, CACD~\cite{chen2014cross}, IMDB~\cite{rothe2018deep}, UTKFace~\cite{zhifei2017cvpr}, AgeDB~\cite{moschoglou2017agedb}, AFAD~\cite{niu2016ordinal}, AAF~\cite{cheng2019exploiting}, FG-NET~\cite{FG_NET}, RFW~\cite{Wang_2019_ICCV}, IMFDB-CVIT~\cite{imfdb}, Asian-DeepGlint~\cite{AsianCeleb}, and PCSO~\cite{deb2017face} are the datasets for training and testing demographic estimation models; and  MS-Celeb-1M~\cite{guo2016ms}, LFW~\cite{huang2008labeled}, IJB-A~\cite{klare2015pushing}, and IJB-C~\cite{maze2018iarpa} are for learning and evaluating face verification models. 
All faces are detected by MTCNN~\cite{zhang2016joint}. 
Each face image is cropped and resized to $112 \times 112$ pixels using a similarity transformation based on the detected landmarks.

\subsection{Implementation Details}
DebFace is trained on a cleaned version of MS-Celeb-1M~\cite{Deng_2019_CVPR}, using the ArcFace architecture~\cite{Deng_2019_CVPR} with $50$ layers for the encoder $E_{Img}$. 
Since there are no demographic labels in MS-Celeb-1M, we first train three demographic attribute estimation models for gender, age, and race, respectively. 
For age estimation, the model is trained on the combination of CACD, IMDB, UTKFace, AgeDB, AFAD, and AAF datasets. 
The gender estimation model is trained on the same datasets except CACD which contains no gender labels. 
We combine AFAD, RFW, IMFDB-CVIT, and PCSO for race estimation training. 
All three models use ResNet~\cite{he2016deep} with $34$ layers for age, $18$ layers for gender and race. 

We predict the demographic labels of MS-Celeb-1M with the well-trained demographic models. 
Our DebFace is then trained on the re-labeled MS-Celeb-1M using SGD with a momentum of $0.9$, a weight decay of $0.01$, and a batch size of $256$. 
The learning rate starts from $0.1$ and drops to $0.0001$ following the schedule at $8$, $13$, and $15$ epochs. 
The dimensionality of the embedding layer of $E_{Img}$ is $4 \times 512$, {\it i.e.}, each attribute representation (gender, age, race, ID) is a $512$-\textit{dim} vector. 
We keep the hyper-parameter setting of AM-Softmax as~\cite{Deng_2019_CVPR}: $s=64$ and $m=0.5$. 
The feature aggregation network $E_{Feat}$ comprises of two linear residual units with P-ReLU and BatchNorm in between. 
$E_{Feat}$ is trained on MS-Celeb-1M by SGD with a learning rate of $0.01$.
The triplet loss margin $\alpha$ is $1.0$. 
The disentangled features of gender, race, and identity are concatenated into a $3 \times 512$-\textit{dim} vector, which inputs to $E_{Feat}$. 
The network is then trained to output a $512$-\textit{dim} representation for face recognition on biased datasets. 
Our source code is available at \url{https://github.com/gongsixue/DebFace.git}.

\subsection{De-biasing Face Verification}
\label{sec:face_verify}
\textbf{Baseline:} We compare DebFace-ID with a regular face representation model which has the same architecture as the shared feature encoder of DebFace. Referred to as BaseFace, this baseline model is also trained on MS-Celeb-1M, with the representation dimension of $512$.  

To show the efficacy of DebFace-ID on bias mitigation, we evaluate the verification performance of DebFace-ID and BaseFace on faces from each demographic cohort.
There are $48$ total cohorts given the combination of demographic attributes including $2$ gender (male, female), $4$ race~\footnote{To clarify, we consider two race groups, Black and White; and two ethnicity groups, East Asian and Indian. The word race denotes both race and ethnicity in this paper.} (Black, White, East Asian, Indian), and $6$ age group ($0-12$, $13-18$, $19-34$, $35-44$, $45-54$, $55-100$). We combine CACD, AgeDB, CVIT, and a subset of Asian-DeepGlint as the testing set. Overlapping identities among these datasets are removed. 
IMDB is excluded from the testing set due to its massive number of wrong ID labels. 
For the dataset without certain demographic labels, we simply use the corresponding models to predict the labels.
We report the  Area Under the Curve (AUC) of the Receiver Operating Characteristics (ROC).
We define the degree of bias, termed {\it biasness}, as the standard deviation of performance across cohorts.

\begin{figure}[t!]
	    \captionsetup{font=footnotesize}
	    \centering
	    \begin{subfigure}[b]{0.498\linewidth}
	    \centering
	    \includegraphics[width=1.0\linewidth]{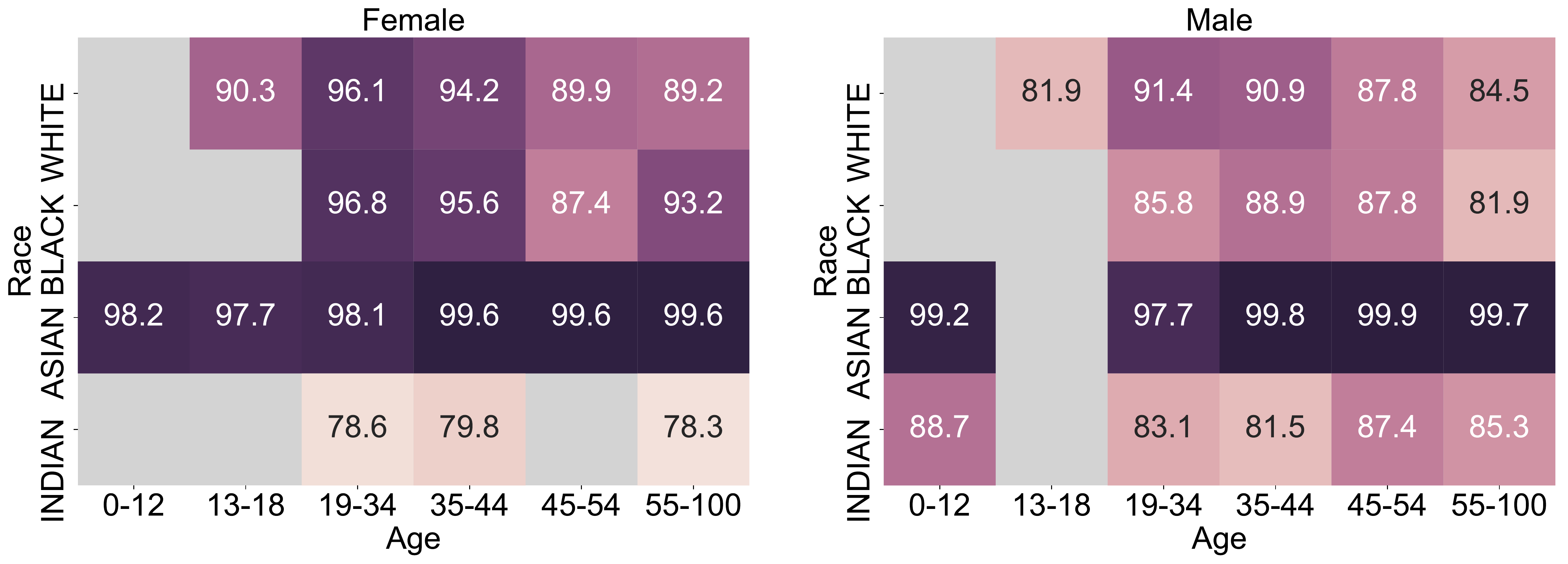}
	    \caption{{\scriptsize BaseFace}}
	    \label{fig:baseface_cell}
	    \end{subfigure}\hfill		
		\begin{subfigure}[b]{0.498\linewidth}
		\centering
		\includegraphics[width=1.0\linewidth]{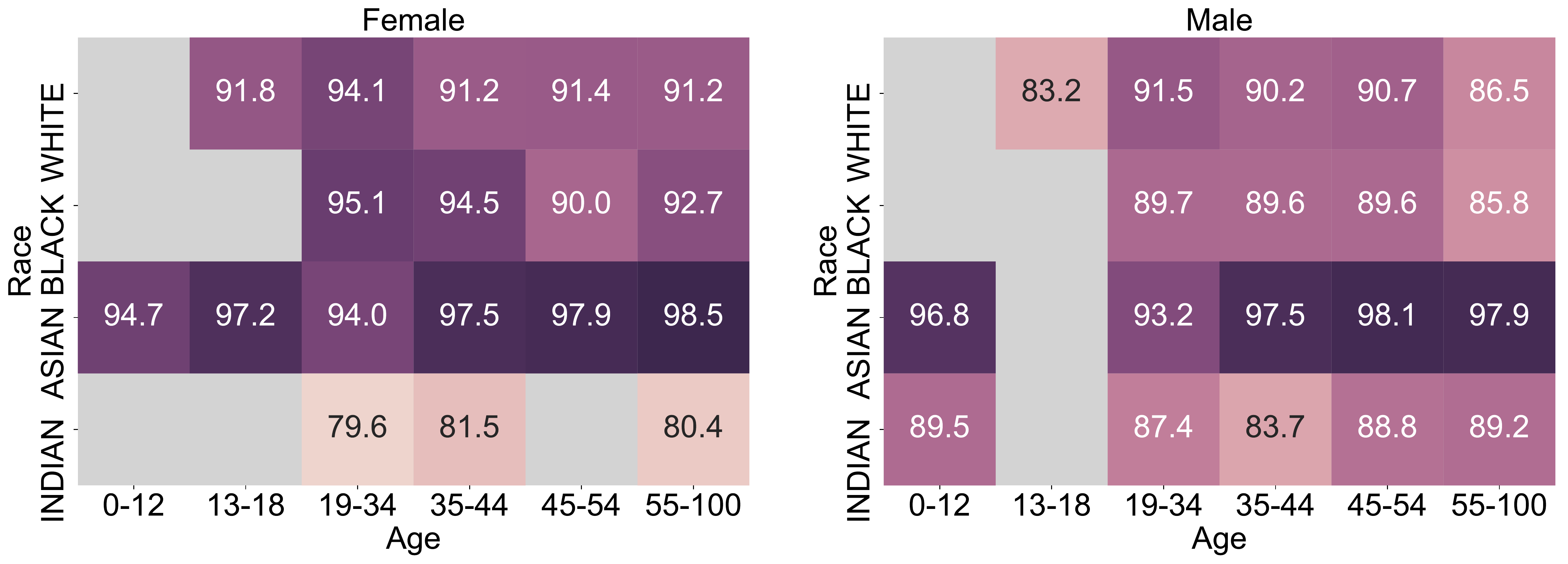}
	    \caption{{\scriptsize DebFace-ID}}
	    \label{fig:debface_cell}
	    \end{subfigure}\\
	    \caption{\footnotesize{Face Verification AUC (\%) on each demographic cohort. The cohorts are chosen based on the three attributes, {\it i.e.}, gender, age, and race. To fit the results into a $2$D plot, we show the performance of male and female separately. Due to the limited number of face images in some cohorts, their results are gray cells.}} 
	    \label{fig:facecell}
\end{figure}

\begin{figure}[t!]
	    \captionsetup{font=footnotesize}
	    \centering
	    \begin{subfigure}[b]{0.315\linewidth}
	    \includegraphics[width=\linewidth]{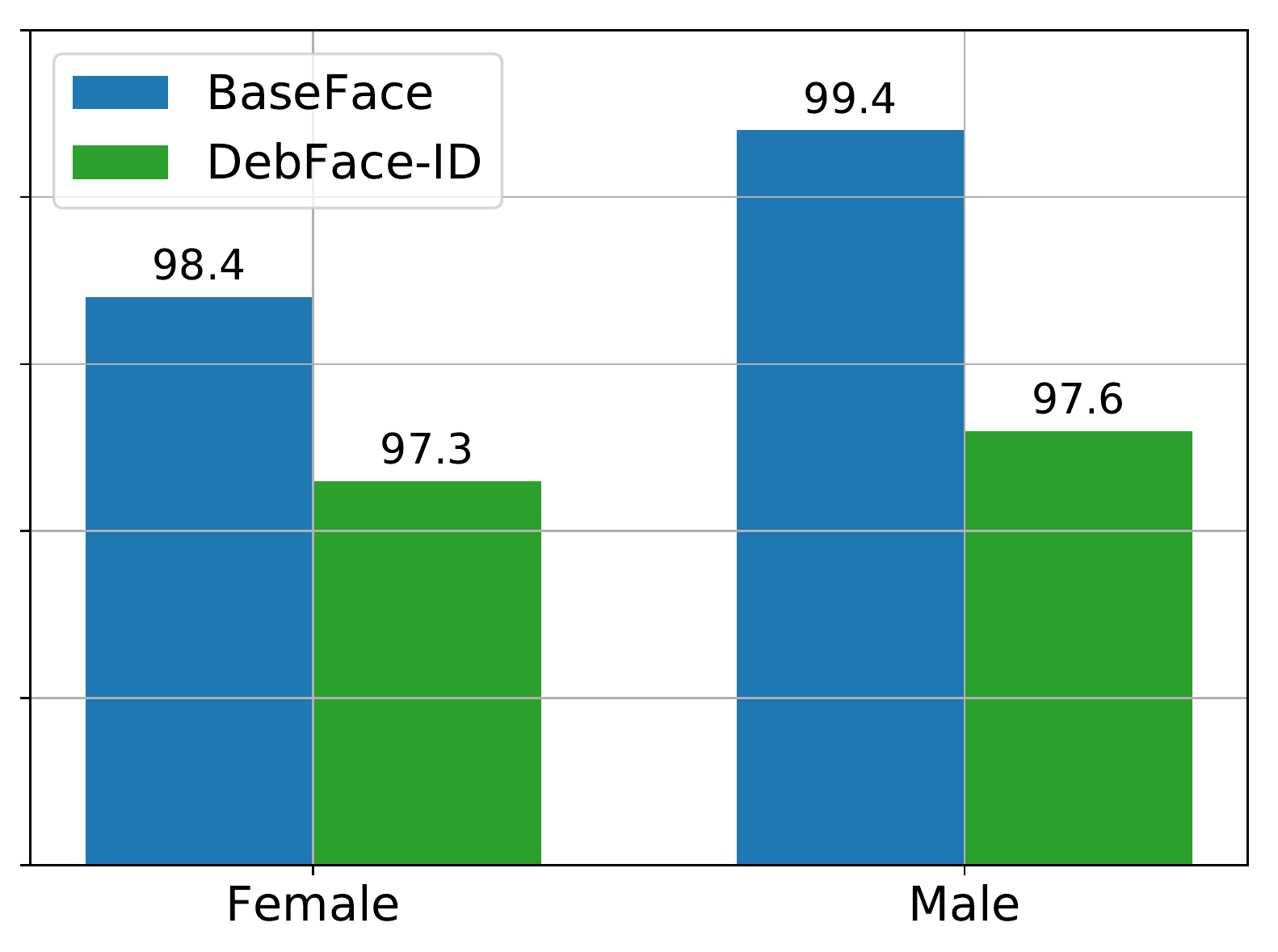}
	    \caption{{\scriptsize Gender}}
	    \label{fig:face_gender_base}
	    \end{subfigure}\hfill		
		\begin{subfigure}[b]{0.36\linewidth}
		\includegraphics[width=\linewidth]{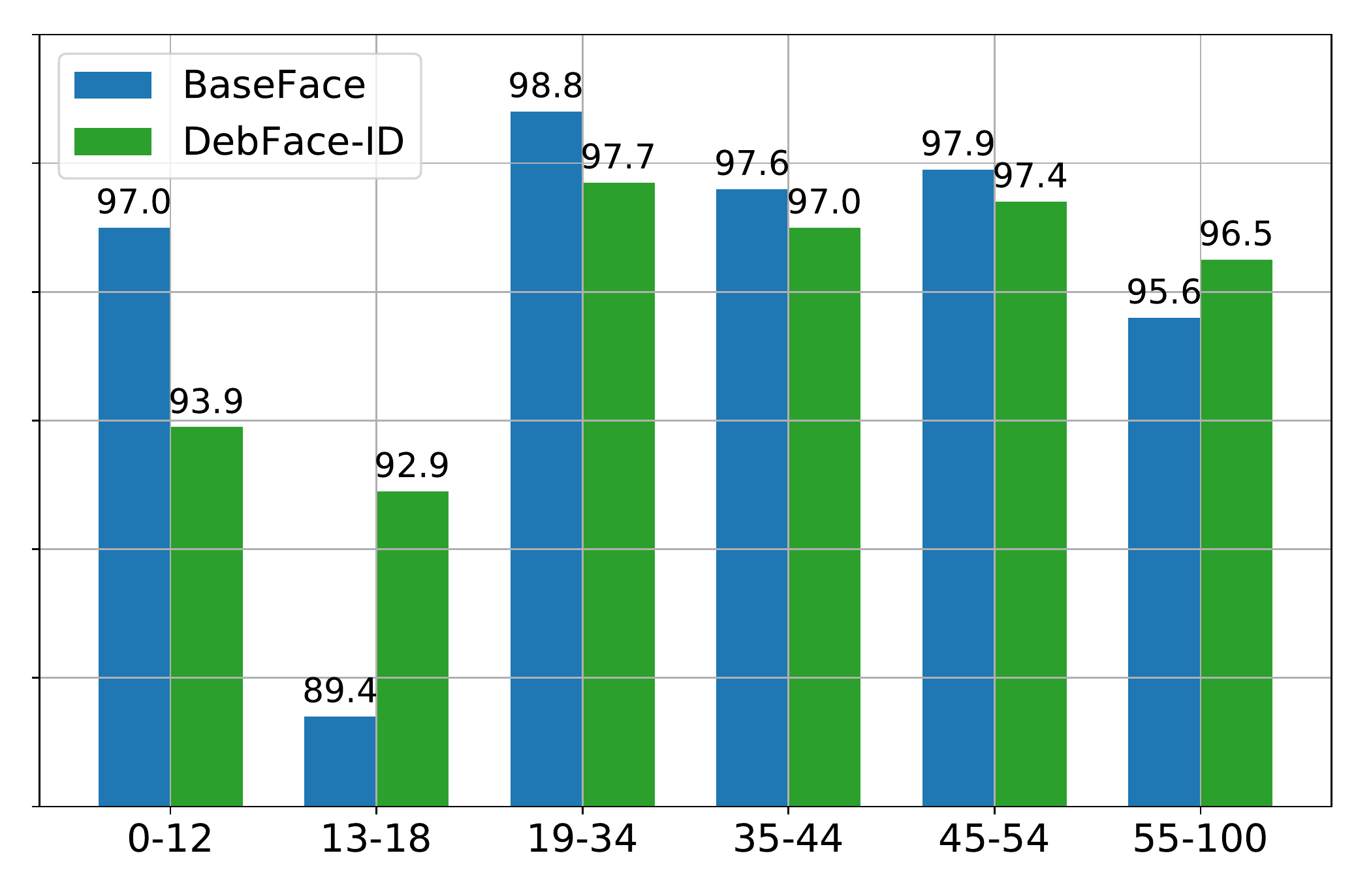}
	    \caption{{\scriptsize Age}}
	    \label{fig:face_age_base}
	    \end{subfigure}\hfill    
	    \begin{subfigure}[b]{0.315\linewidth}
	    \includegraphics[width=\linewidth]{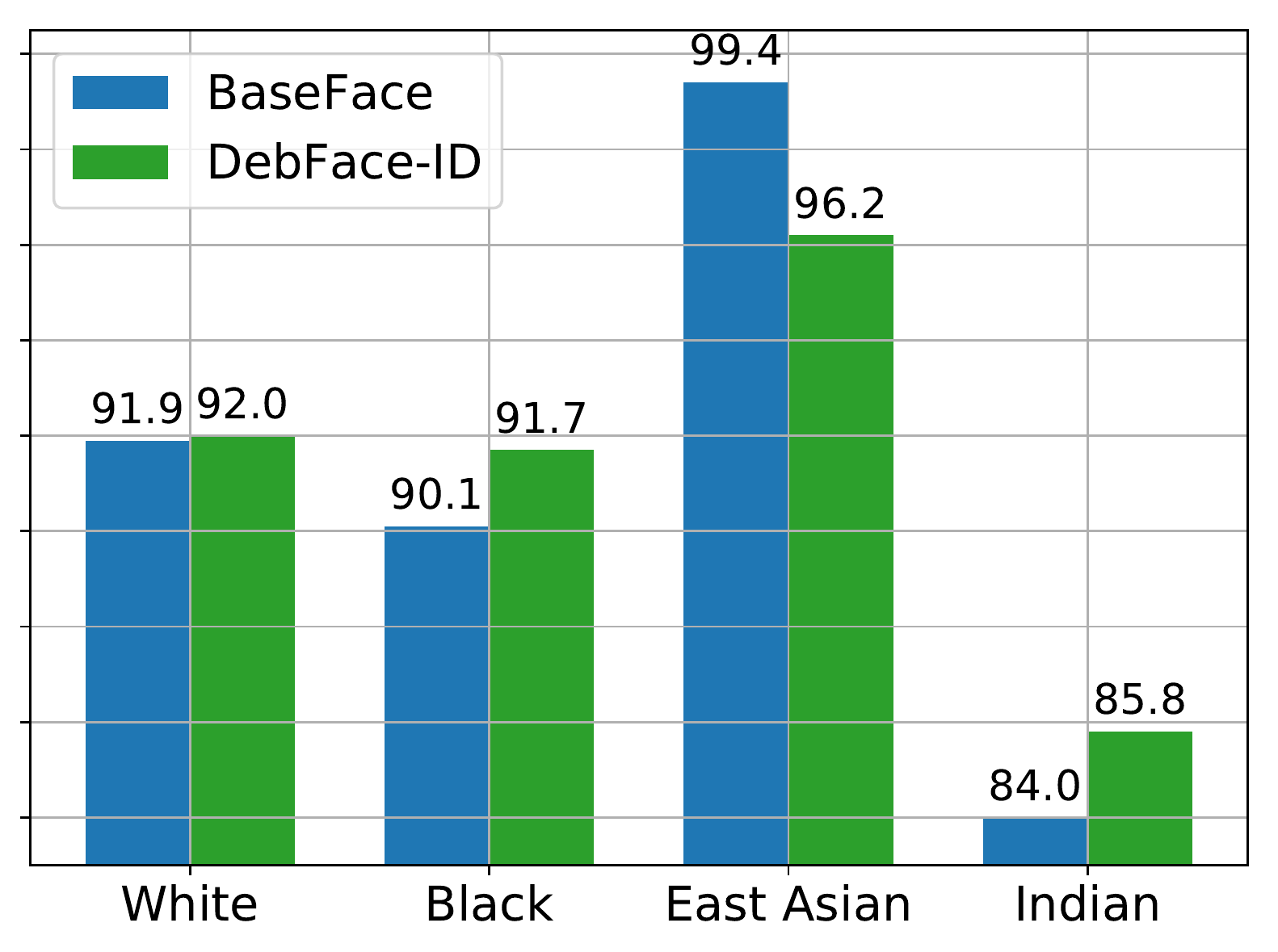}
	    \caption{{\scriptsize Race}}
	    \label{fig:face_race_base}
	    \end{subfigure}\\
	    \caption{\footnotesize{The overall performance of face verification AUC (\%) on gender, age, and race.}}
	    \label{fig:face_bar_base}
\end{figure}

Figure~\ref{fig:facecell} shows the face verification results of BaseFace and DebFace-ID on each cohort. 
That is, for a particular face representation ({\it e.g.}, DebFace-ID), we report its AUC on each cohort by putting the number in the corresponding cell. 
From these heatmaps, we observe that both DebFace-ID and BaseFace present bias in face verification, where the performance on some cohorts are significantly worse, especially the cohorts of Indian female and elderly people.
Compared to BaseFace, DebFace-ID suggests less bias and the difference of AUC is smaller, where the heatmap exhibits smoother edges. Figure~\ref{fig:face_bar_base} shows the performance of face verification on $12$ demographic cohorts.
Both DebFace-ID and BaseFace present similar relative accuracies across cohorts.
For example, both algorithms perform worse on the younger age cohorts than on adults; and the performance on the Indian is significantly lower than on the other races.
DebFace-ID decreases the bias by gaining discriminative face features for cohorts with less images in spite of the reduction in the performance on cohorts with more samples.

\subsection{De-biasing Demographic Attribute Estimation}
\textbf{Baseline:} We further explore the bias of demographic attribute estimation and compare demographic attribute classifiers of DebFace with baseline estimation models. We train three demographic estimation models, namely, gender estimation (BaseGender), age estimation (BaseAge), and race estimation (BaseRace), on the same training set as DebFace. 
For fairness, all three models have the same architecture as the shared layers of DebFace. 

\begin{figure}[t!]
	    \captionsetup{font=footnotesize}
	    \centering
	    \begin{adjustbox}{minipage=\linewidth,scale=1}
	    \begin{subfigure}[b]{0.2\linewidth}
	    \centering
	    \includegraphics[width=\linewidth]{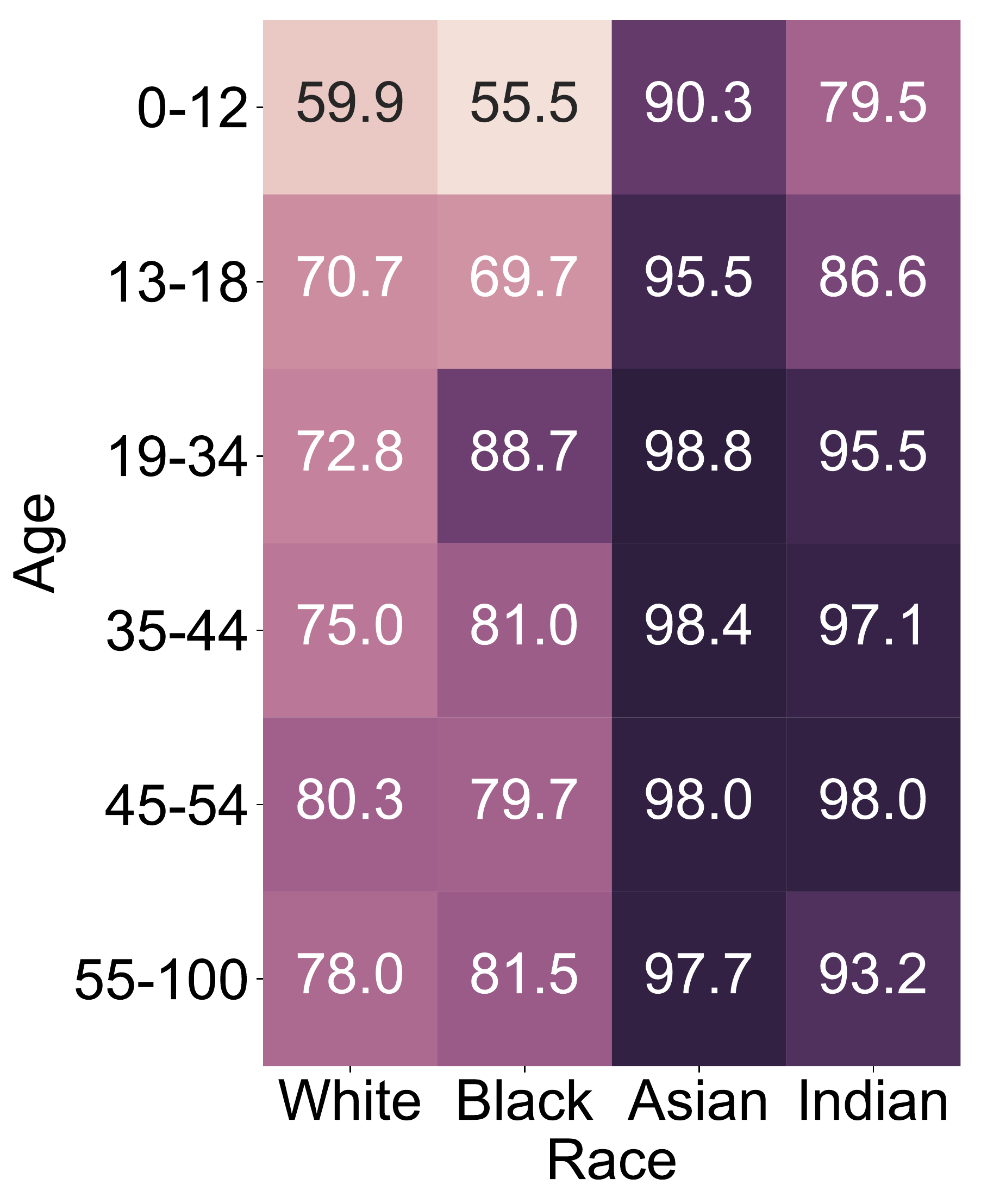}
	    \caption{{\scriptsize BaseGender}}
	    \label{fig:gender_base}
	    \end{subfigure}\hfill
	    \begin{subfigure}[b]{0.2\linewidth}
	    \centering
		\includegraphics[width=\linewidth]{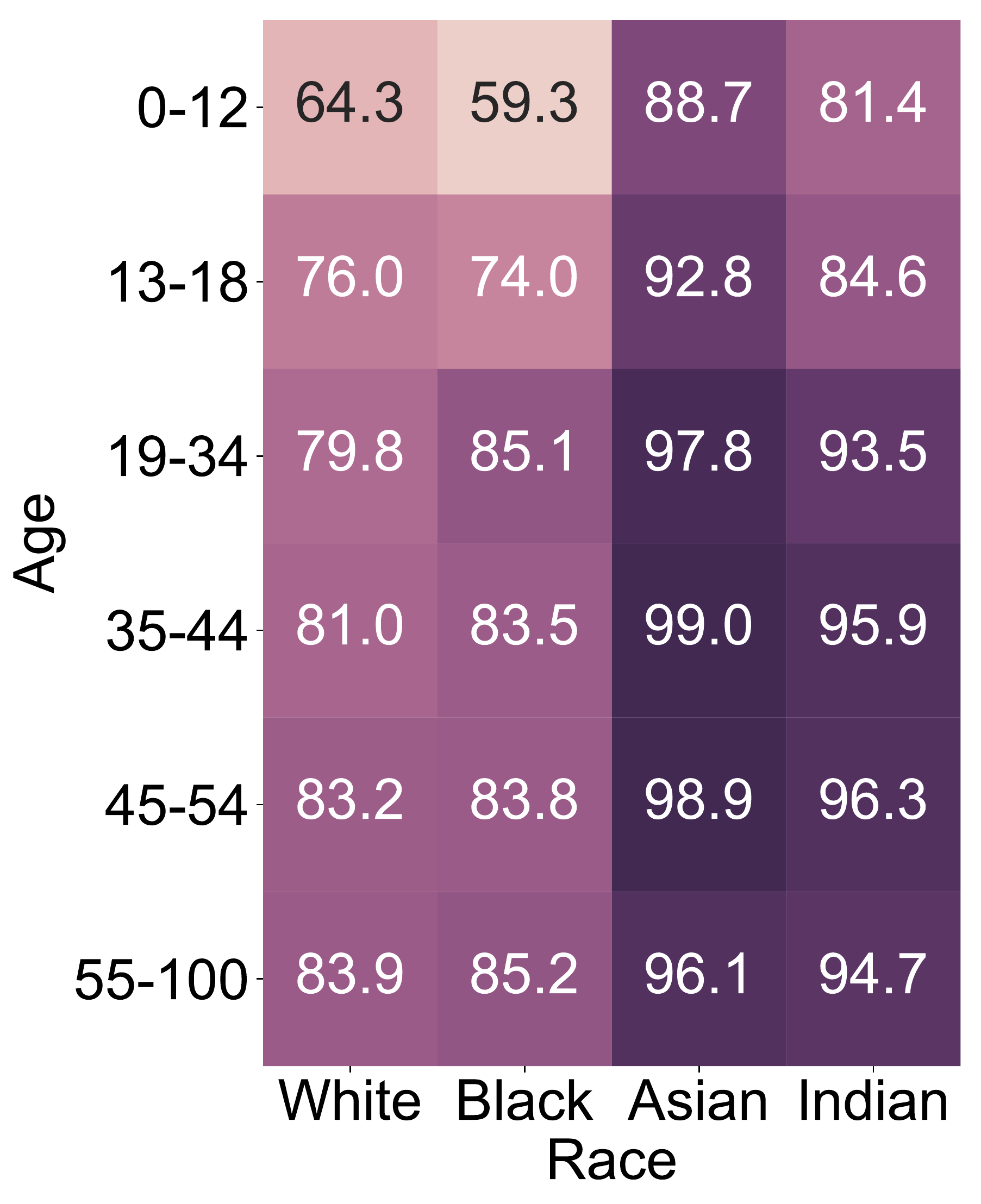}
	    \caption{{\scriptsize DebFace-G}}
	    \label{fig:gender_deb}
	    \end{subfigure}\hfill
	     \begin{subfigure}[b]{0.15\linewidth}
	     \centering
	    \includegraphics[width=0.93\linewidth]{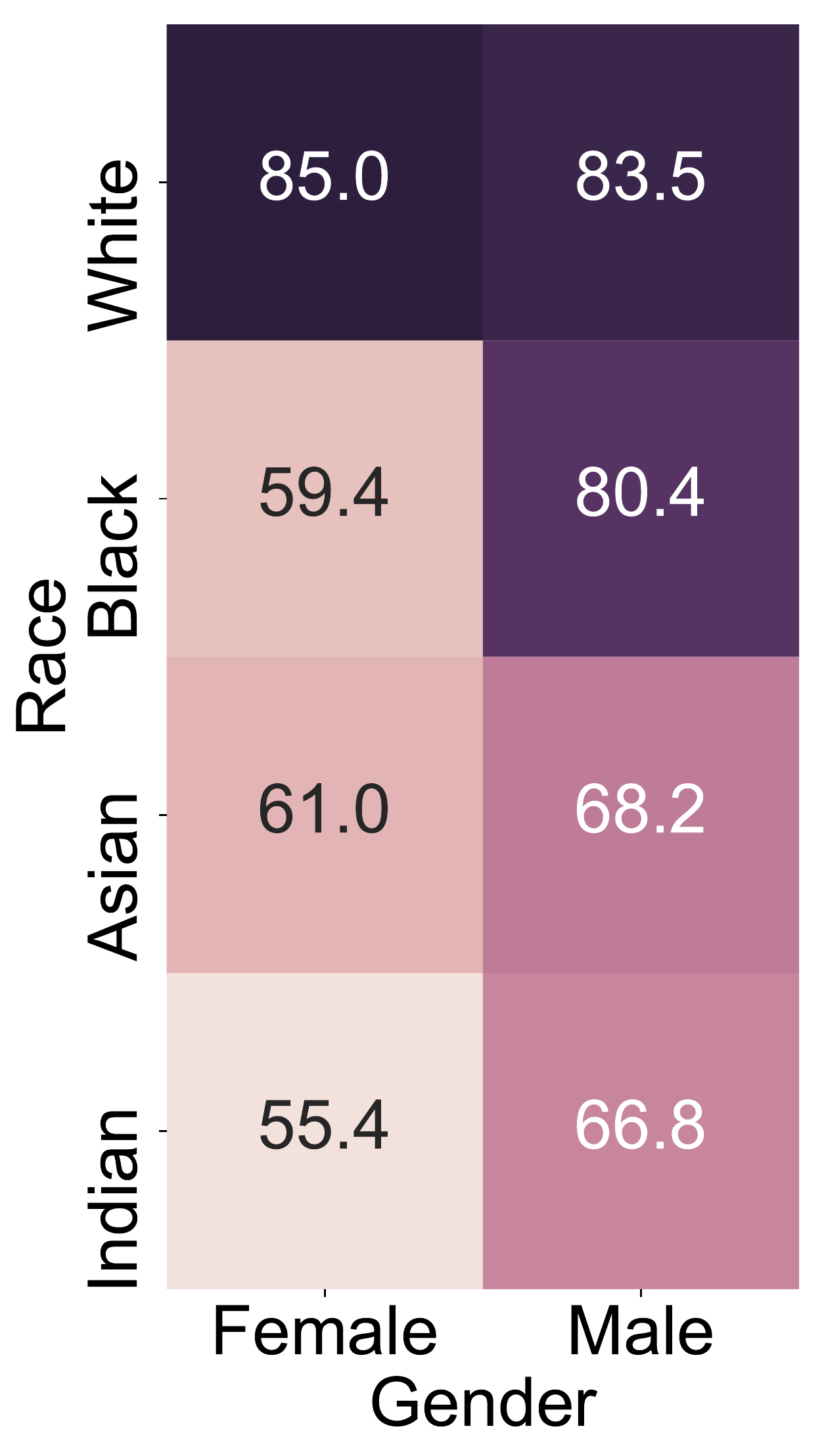}
	    \caption{{\scriptsize BaseAge}}
	    \label{fig:age_base}
	    \end{subfigure}\hfill
	    \begin{subfigure}[b]{0.15\linewidth}
	    \centering
	    \includegraphics[width=0.93\linewidth]{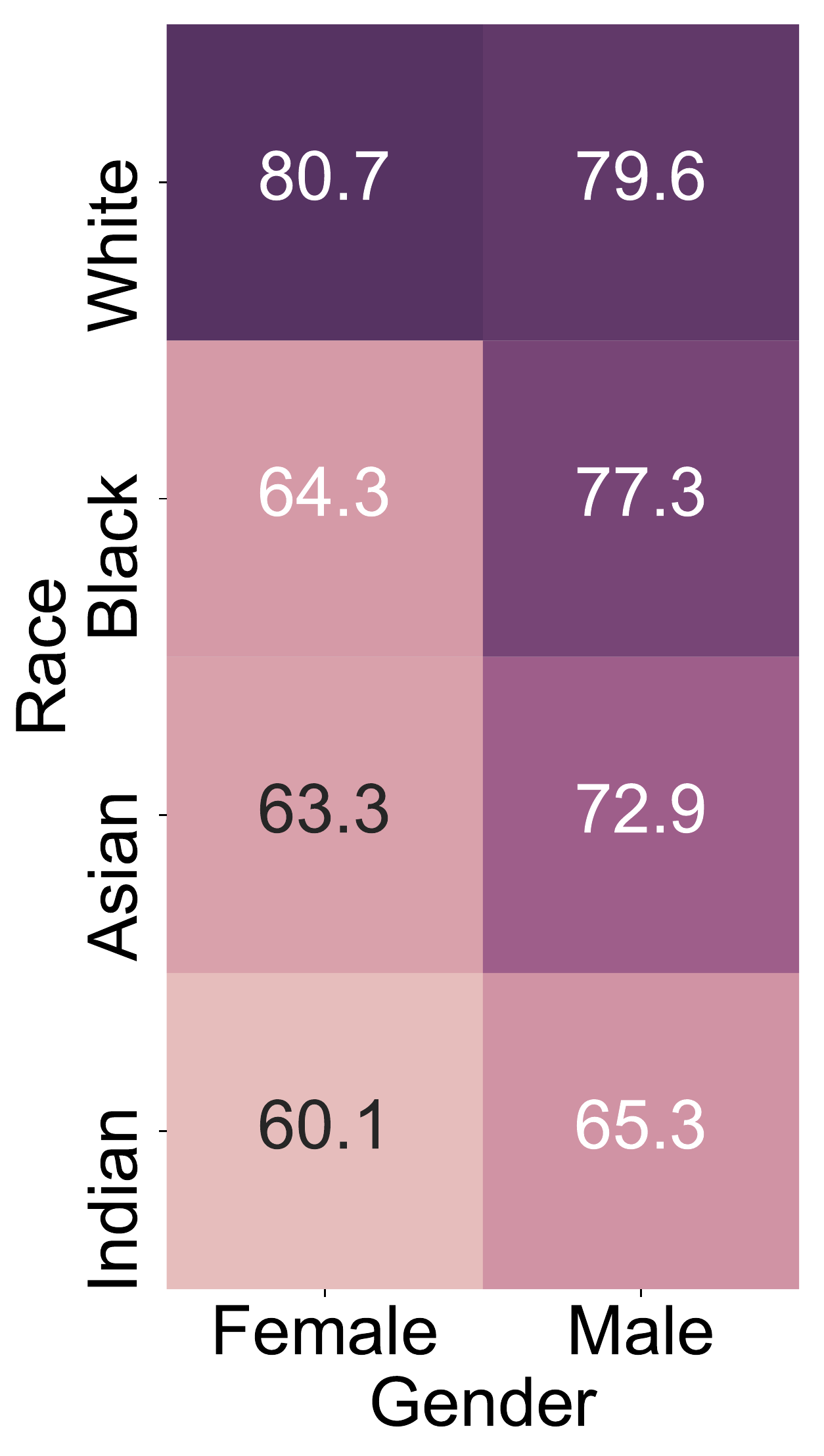}
	    \caption{{\scriptsize DebFace-A}}
	    \label{fig:age_deb}
	    \end{subfigure}\hfill
	    \begin{subfigure}[b]{0.15\linewidth}
	    \centering
		\includegraphics[width=0.86\linewidth]{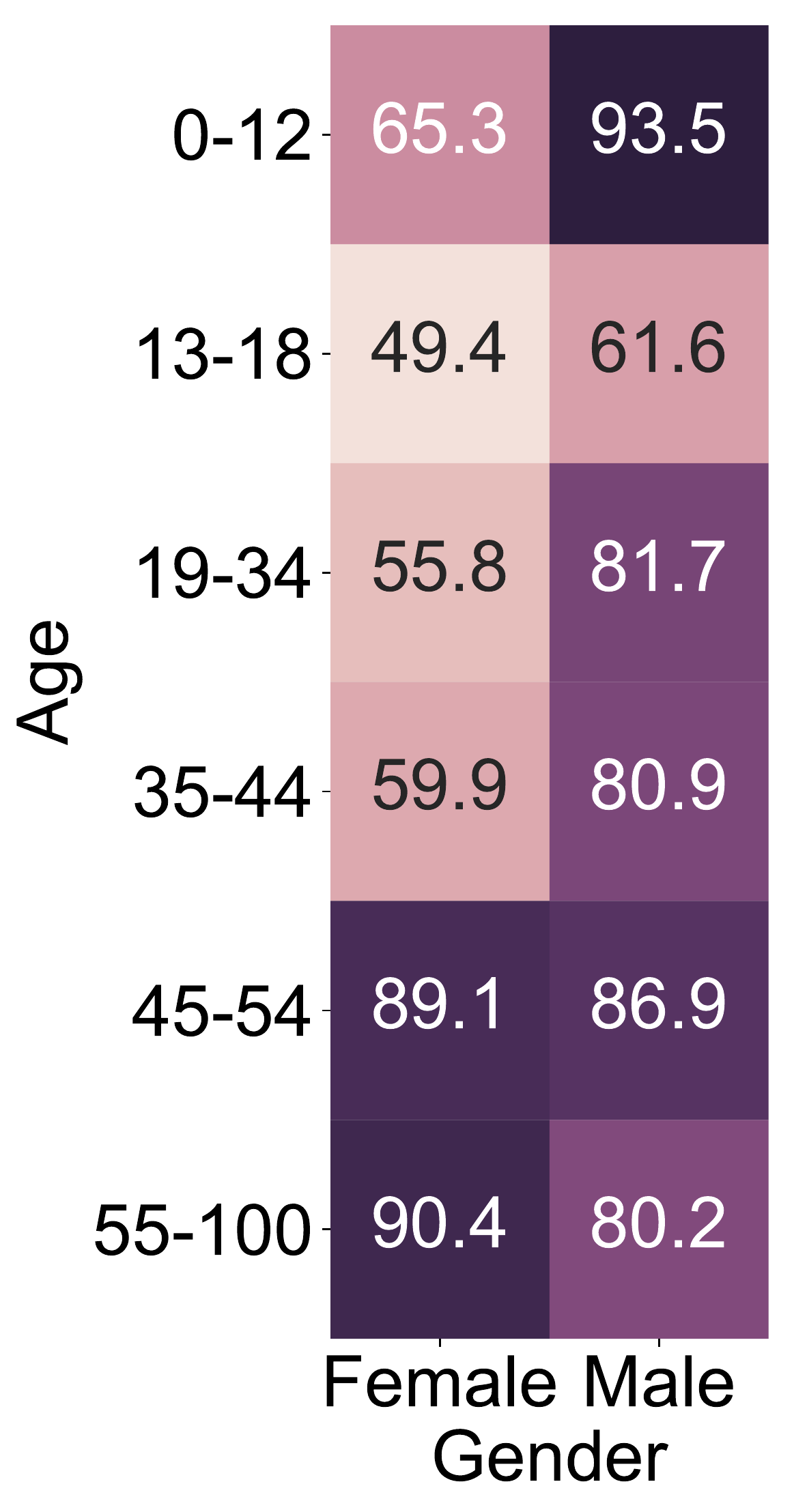}
	    \caption{{\scriptsize BaseRace}}
	    \label{fig:race_base}
	    \end{subfigure}\hfill
	    \begin{subfigure}[b]{0.15\linewidth}
	    \centering
	    \includegraphics[width=0.86\linewidth]{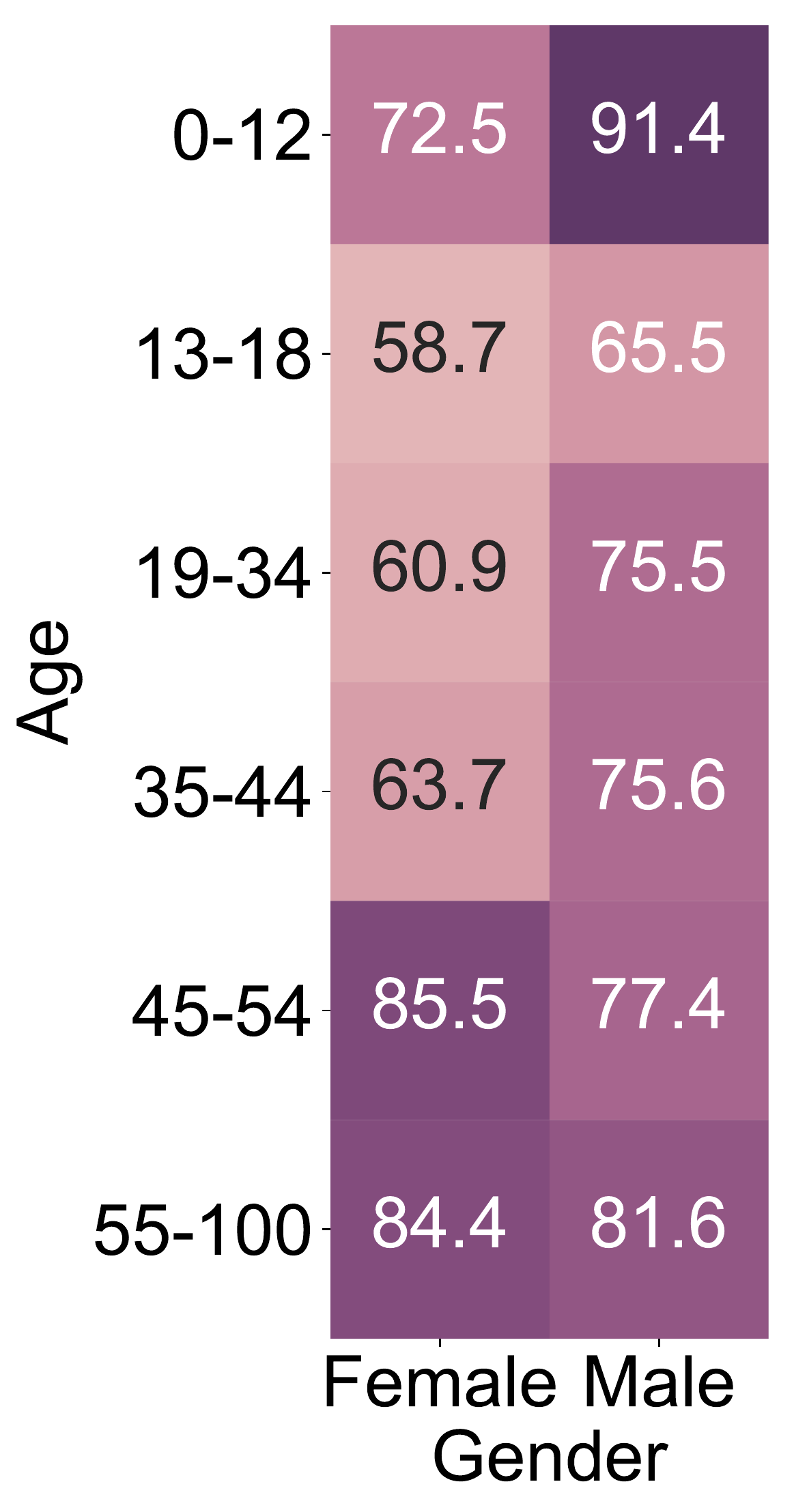}
	    \caption{{\scriptsize DebFace-R}}
	    \label{fig:race_deb}
	    \end{subfigure}\\
	    \end{adjustbox}
	    \caption{\footnotesize{Classification accuracy (\%) of demographic attribute estimations on faces of different cohorts, by DebFace and the baselines. For simplicity, we use DebFace-G, DebFace-A, and DebFace-R to represent the gender, age, and race classifier of DebFace.}}
	    \label{fig:demog_bias}
\end{figure}

We combine the four datasets mentioned in Sec.~\ref{sec:face_verify} with IMDB as the global testing set.
As all demographic estimations are treated as classification problems, the classification accuracy is used as the performance metric. 
As shown in Fig.~\ref{fig:demog_bias}, all demographic attribute estimations present significant bias. 
For gender estimation, both algorithms perform worse on the White and Black cohorts than on East Asian and Indian. 
In addition, the performance on young children is significantly worse than on adults. 
In general, the race estimation models perform better on the male cohort than on female. 
Compared to gender, race estimation shows higher bias in terms of age. 
Both baseline methods and DebFace perform worse on cohorts in age between $13$ to $44$ than in other age groups.

\begin{table}[t!]
    \centering
    \captionsetup{font=footnotesize}
    \caption{\footnotesize{Biasness of Face Recognition and Demographic Attribute Estimation.}} 
    \label{table_biasness}
    \scalebox{0.85}{
    \begin{tabular}{P{1.2cm}P{0.2cm} P{1.2cm} P{1.2cm} P{1.2cm} P{1.2cm}P{0.2cm} P{1.2cm} P{1.2cm} P{1.2cm}}
        \toprule
        \multirow{2}{*}{Method} && \multicolumn{4}{c}{Face Verification} && \multicolumn{3}{c}{Demographic Estimation} \\
        \cline{3-6} \cline{8-10}
        && All & Gender & Age & Race && Gender & Age & Race \\
        \midrule
        Baseline && 6.83 & 0.50 & 3.13 & 5.49 && 12.38 & 10.83 & 14.58 \\
        DebFace && \textbf{5.07} & \textbf{0.15} & \textbf{1.83} & \textbf{3.70} && \textbf{10.22} & \textbf{7.61} & \textbf{10.00} \\
        \bottomrule
    \end{tabular}}
\end{table}

Similar to race, age estimation still achieves better performance on male than on female. 
Moreover, the white cohort shows dominant advantages over other races in age estimation. 
In spite of the existing bias in demographic attribute estimations, the proposed DebFace is still able to mitigate bias derived from algorithms. Compared to Fig.~\ref{fig:gender_base}, ~\ref{fig:race_base}, ~\ref{fig:age_base}, cells in Fig.~\ref{fig:gender_deb}, ~\ref{fig:race_deb}, ~\ref{fig:age_deb} present more uniform colors.
We summarize the biasness of DebFace and baseline models for both face recognition and demographic attribute estimations in Tab.~\ref{table_biasness}. In general, we observe DebFace substantially reduces biasness for both tasks. For the task with larger biasness, the reduction of biasness is larger.

  \begin{figure}[t!]
	    \captionsetup{font=footnotesize}
	    \centering
	    \begin{subfigure}[b]{0.15\linewidth}
	    \includegraphics[width=\linewidth]{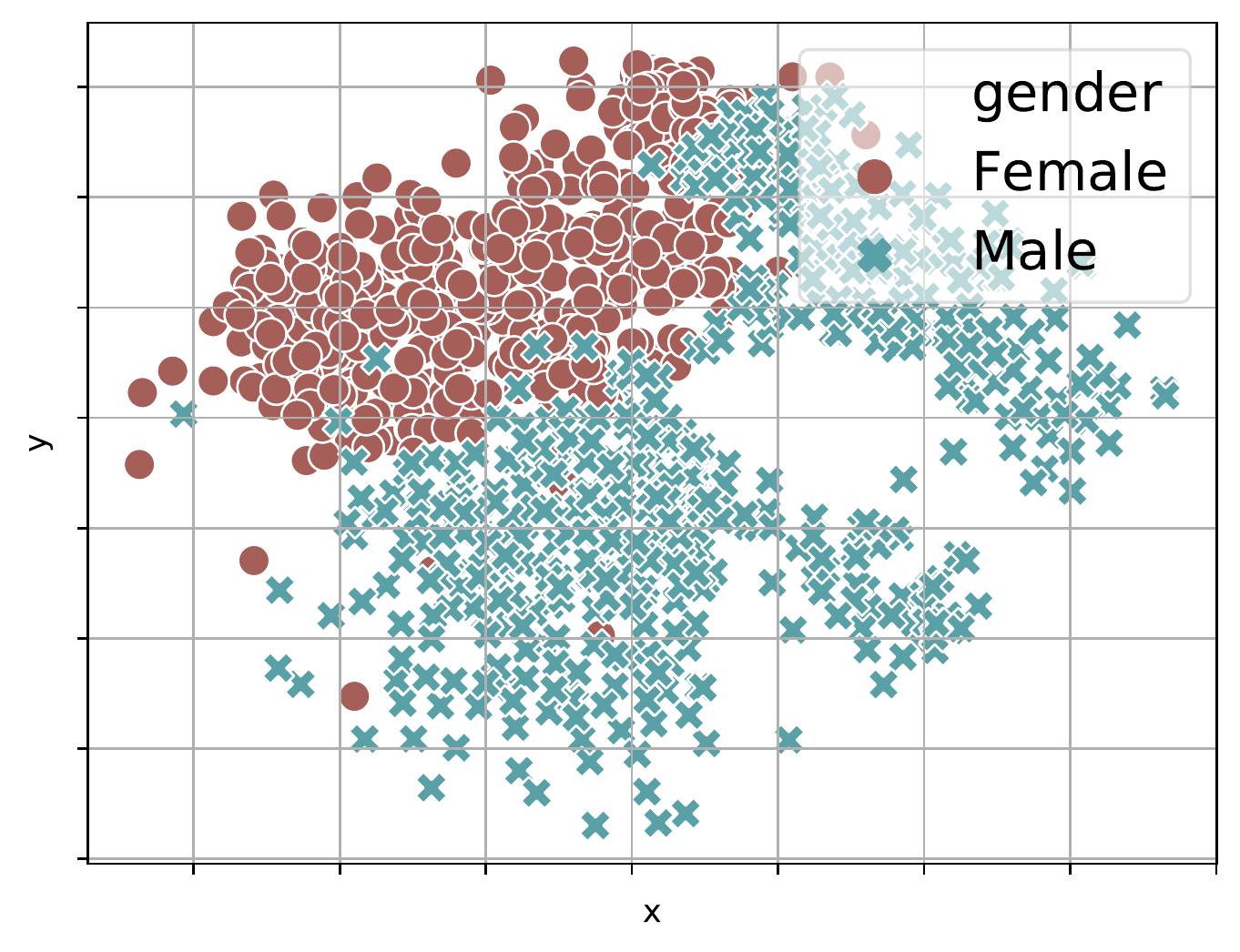}
	    \caption{{\scriptsize BaseFace}}
	    \label{fig:tsne_gender_base}
	    \end{subfigure}\hfill
	    \begin{subfigure}[b]{0.16\linewidth}
		\includegraphics[width=0.94\linewidth]{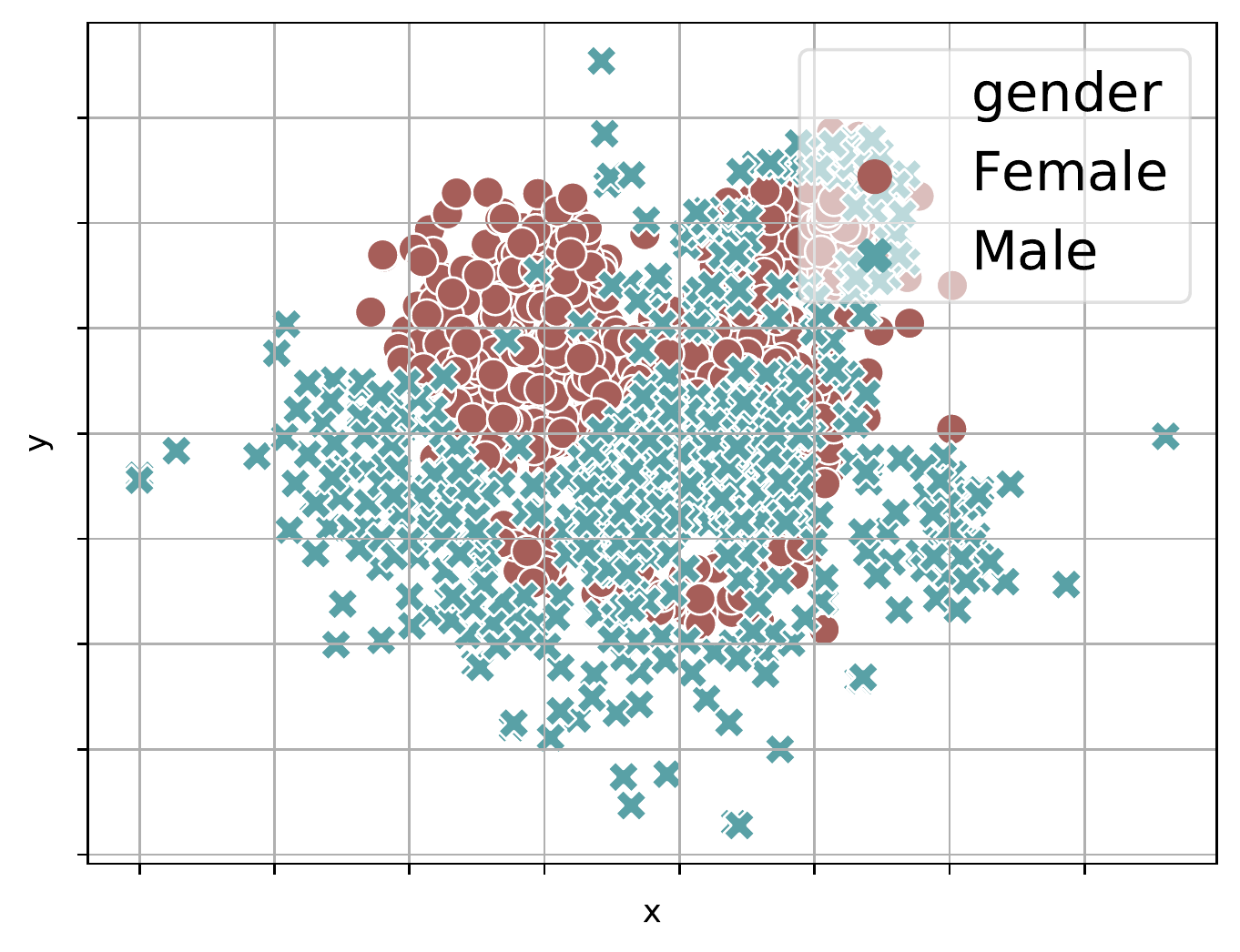}
	    \caption{{\scriptsize DebFace-ID}}
	    \label{fig:tsne_gender_deb}
	    \end{subfigure}\hfill
	     \begin{subfigure}[b]{0.15\linewidth}
	    \includegraphics[width=\linewidth]{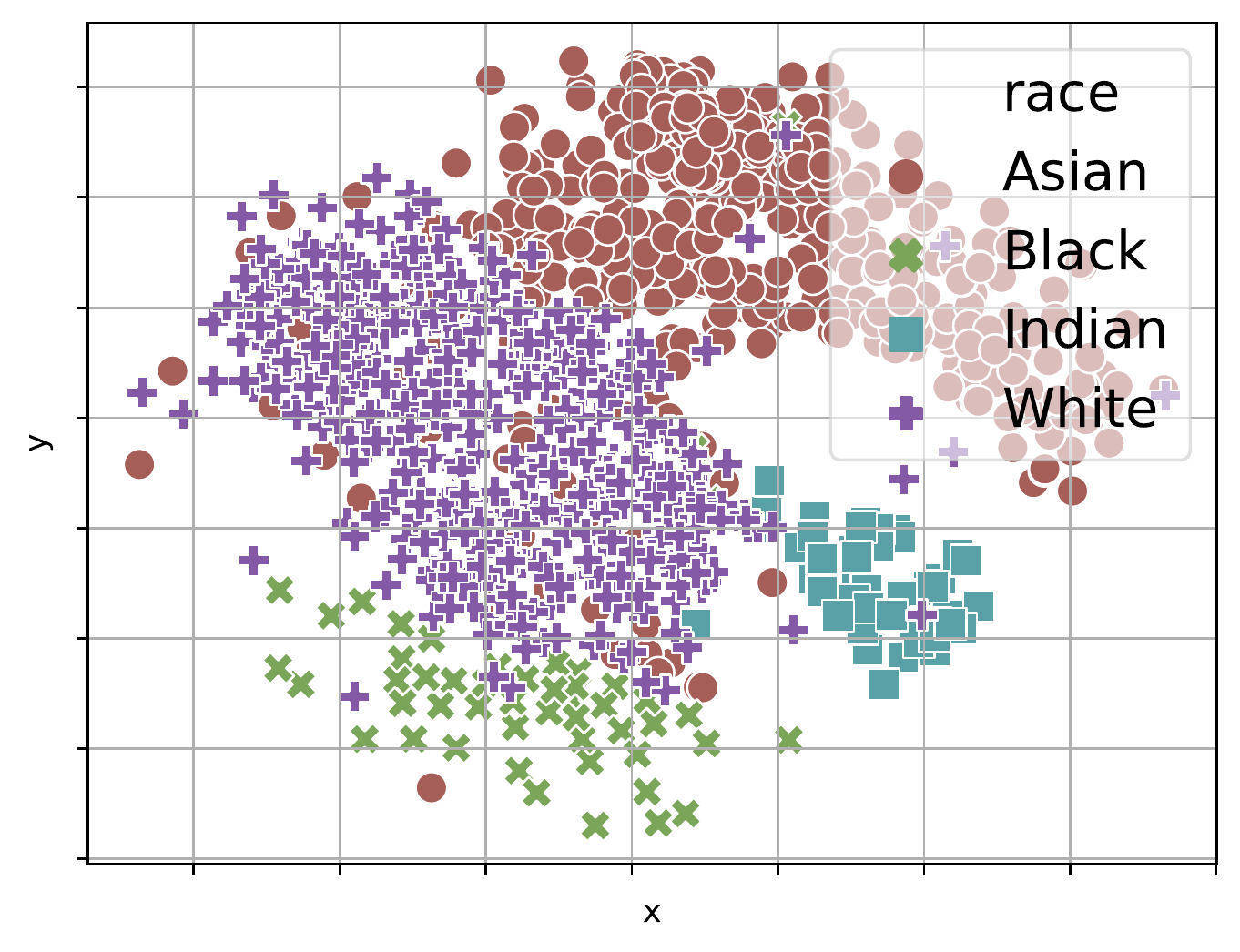}
	    \caption{{\scriptsize BaseFace}}
	    \label{fig:tsne_race_base}
	    \end{subfigure}\hfill
	    \begin{subfigure}[b]{0.16\linewidth}
	    \includegraphics[width=0.94\linewidth]{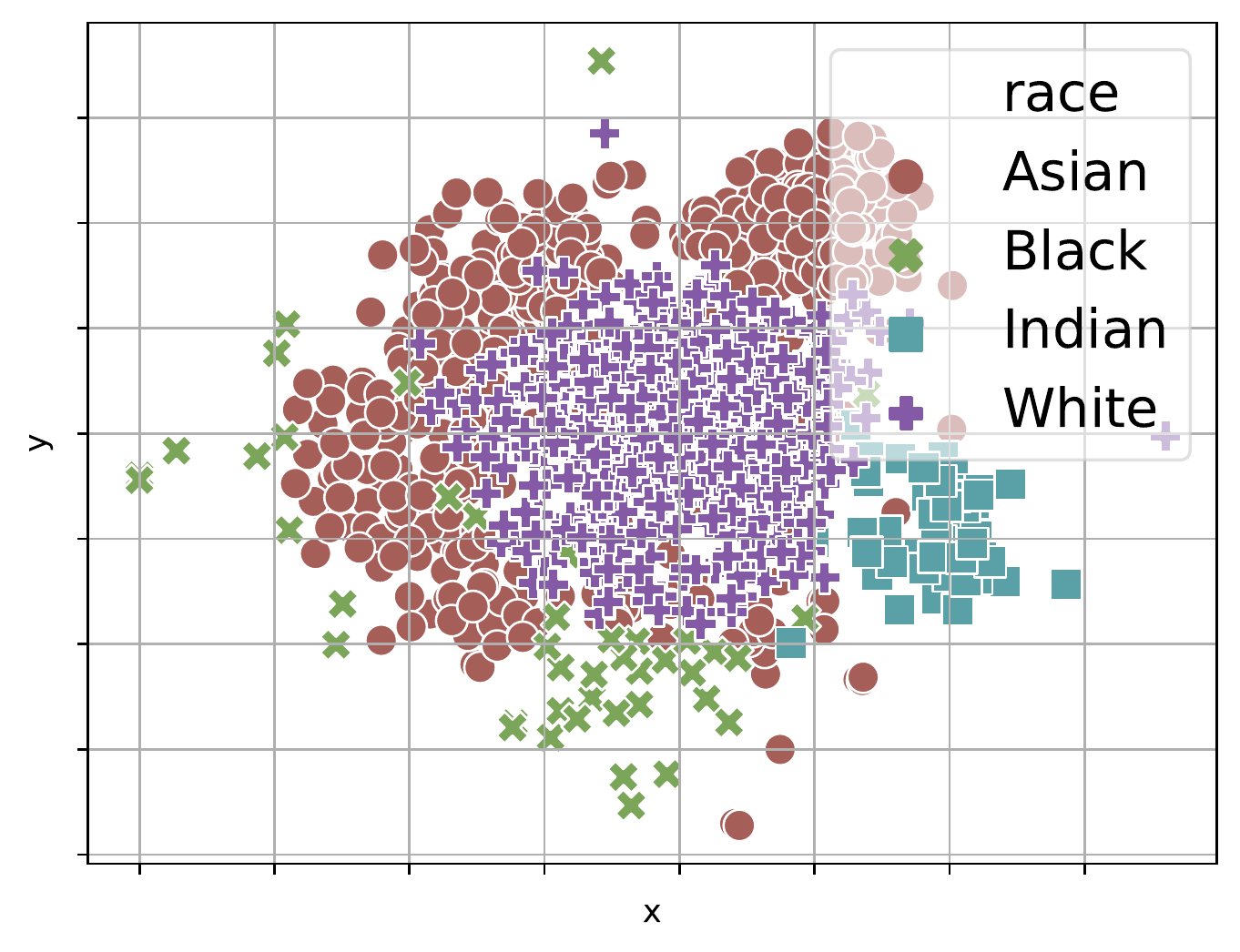}
	    \caption{{\scriptsize DebFace-ID}}
	    \label{fig:tsne_race_deb}
	    \end{subfigure}\hfill
	    \begin{subfigure}[b]{0.15\linewidth}
		\includegraphics[width=\linewidth]{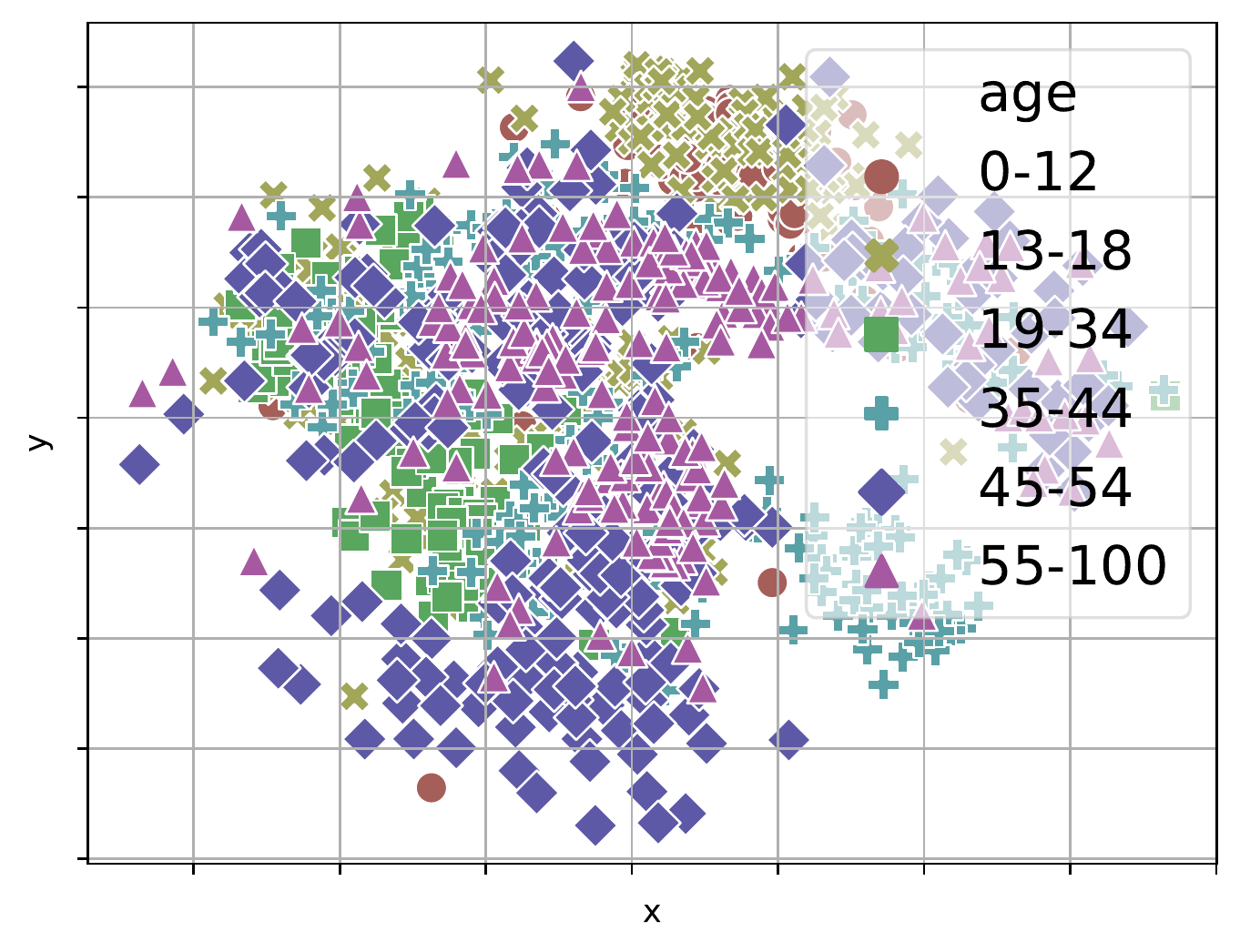}
	    \caption{{\scriptsize BaseFace}}
	    \label{fig:tsne_age_base}
	    \end{subfigure}\hfill
	    \begin{subfigure}[b]{0.16\linewidth}
	    \includegraphics[width=0.94\linewidth]{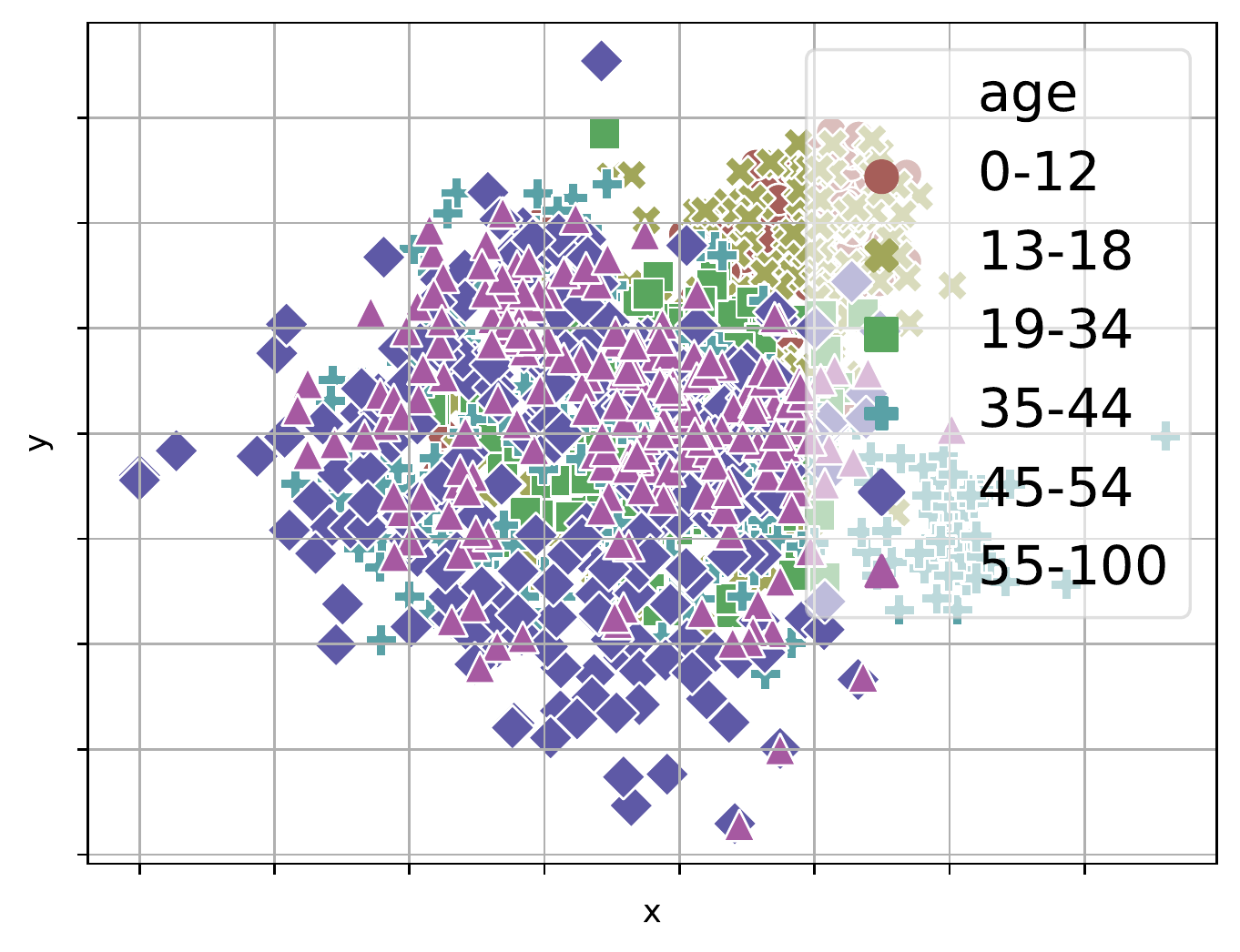}
	    \caption{{\scriptsize DebFace-ID}}
	    \label{fig:tsne_age_deb}
	    \end{subfigure}\\
	    \caption{\footnotesize{The distribution of face identity representations of BaseFace and DebFace. Both collections of feature vectors are extracted from images of the same dataset. Different colors and shapes represent different demographic attributes. Zoom in for details.}}
	    \label{fig:tsne_face}
\end{figure}
\begin{figure}[t!]
	    \captionsetup{font=footnotesize}
	    \centering
	    \begin{subfigure}[b]{\linewidth}
		\centering
		\begin{minipage}[c]{0.08\textwidth}
		\tiny Original
	    \end{minipage}\hfill
	    \begin{minipage}[c]{0.92\textwidth}
		\includegraphics[width=0.062\linewidth]{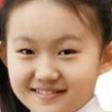}\hfill
		\includegraphics[width=0.062\linewidth]{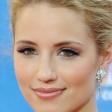}\hfill
		\includegraphics[width=0.062\linewidth]{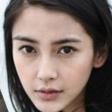}\hfill
		\includegraphics[width=0.062\linewidth]{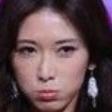}\hfill
		\includegraphics[width=0.062\linewidth]{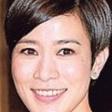}\hfill
		\includegraphics[width=0.062\linewidth]{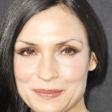}\hfill
		\includegraphics[width=0.062\linewidth]{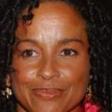}\hfill
		\includegraphics[width=0.062\linewidth]{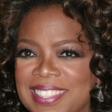}\hfill
		\includegraphics[width=0.062\linewidth]{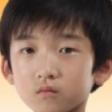}\hfill
		\includegraphics[width=0.062\linewidth]{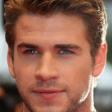}\hfill
		\includegraphics[width=0.062\linewidth]{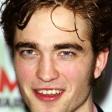}\hfill
		\includegraphics[width=0.062\linewidth]{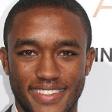}\hfill
		\includegraphics[width=0.062\linewidth]{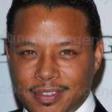}\hfill
		\includegraphics[width=0.062\linewidth]{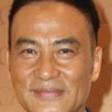}\hfill
		\includegraphics[width=0.062\linewidth]{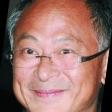}\hfill
		\includegraphics[width=0.062\linewidth]{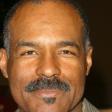}
		\end{minipage}
	    \end{subfigure}\\
	    
	    \begin{subfigure}[b]{\linewidth}
		\centering
		\begin{minipage}[c]{0.08\textwidth}
		\tiny{BaseFace}
	    \end{minipage}\hfill
	    \begin{minipage}[c]{0.92\textwidth}
		\includegraphics[width=0.062\linewidth]{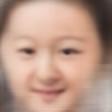}\hfill
		\includegraphics[width=0.062\linewidth]{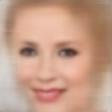}\hfill
		\includegraphics[width=0.062\linewidth]{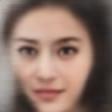}\hfill
		\includegraphics[width=0.062\linewidth]{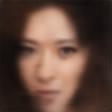}\hfill
		\includegraphics[width=0.062\linewidth]{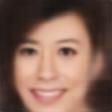}\hfill
		\includegraphics[width=0.062\linewidth]{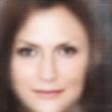}\hfill
		\includegraphics[width=0.062\linewidth]{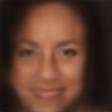}\hfill
		\includegraphics[width=0.062\linewidth]{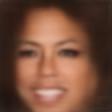}\hfill
		\includegraphics[width=0.062\linewidth]{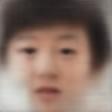}\hfill
		\includegraphics[width=0.062\linewidth]{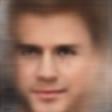}\hfill
		\includegraphics[width=0.062\linewidth]{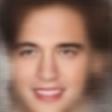}\hfill
		\includegraphics[width=0.062\linewidth]{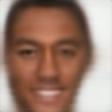}\hfill
		\includegraphics[width=0.062\linewidth]{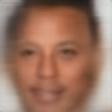}\hfill
		\includegraphics[width=0.062\linewidth]{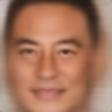}\hfill
		\includegraphics[width=0.062\linewidth]{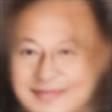}\hfill
		\includegraphics[width=0.062\linewidth]{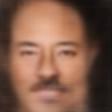}
		\end{minipage}
	    \end{subfigure}\\
	    
	    \begin{subfigure}[b]{\linewidth}
		\centering
		\begin{minipage}[c]{0.08\textwidth}
		\tiny{DebFace-ID}
	    \end{minipage}\hfill
	    \begin{minipage}[c]{0.92\textwidth}
		\includegraphics[width=0.062\linewidth]{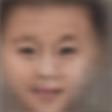}\hfill
		\includegraphics[width=0.062\linewidth]{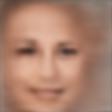}\hfill
		\includegraphics[width=0.062\linewidth]{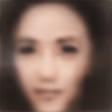}\hfill
		\includegraphics[width=0.062\linewidth]{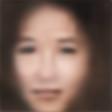}\hfill
		\includegraphics[width=0.062\linewidth]{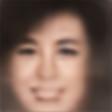}\hfill
		\includegraphics[width=0.062\linewidth]{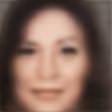}\hfill
		\includegraphics[width=0.062\linewidth]{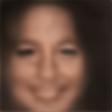}\hfill
		\includegraphics[width=0.062\linewidth]{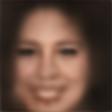}\hfill
		\includegraphics[width=0.062\linewidth]{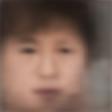}\hfill
		\includegraphics[width=0.062\linewidth]{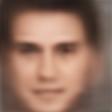}\hfill
		\includegraphics[width=0.062\linewidth]{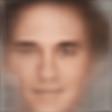}\hfill
		\includegraphics[width=0.062\linewidth]{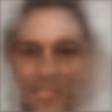}\hfill
		\includegraphics[width=0.062\linewidth]{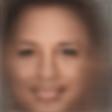}\hfill
		\includegraphics[width=0.062\linewidth]{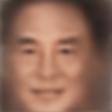}\hfill
		\includegraphics[width=0.062\linewidth]{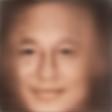}\hfill
		\includegraphics[width=0.062\linewidth]{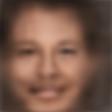}
		\end{minipage}
	    \end{subfigure}\\

        \begin{subfigure}[b]{\linewidth}
		\centering
		\begin{minipage}[c]{0.08\textwidth}
		\tiny{DebFace-G}
	    \end{minipage}\hfill
	    \begin{minipage}[c]{0.92\textwidth}
		\includegraphics[width=0.062\linewidth]{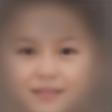}\hfill
		\includegraphics[width=0.062\linewidth]{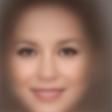}\hfill
		\includegraphics[width=0.062\linewidth]{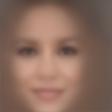}\hfill
		\includegraphics[width=0.062\linewidth]{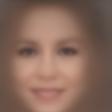}\hfill
		\includegraphics[width=0.062\linewidth]{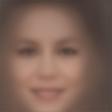}\hfill
		\includegraphics[width=0.062\linewidth]{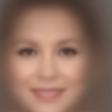}\hfill
		\includegraphics[width=0.062\linewidth]{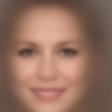}\hfill
		\includegraphics[width=0.062\linewidth]{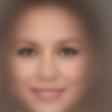}\hfill
		\includegraphics[width=0.062\linewidth]{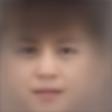}\hfill
		\includegraphics[width=0.062\linewidth]{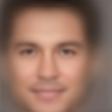}\hfill
		\includegraphics[width=0.062\linewidth]{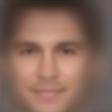}\hfill
		\includegraphics[width=0.062\linewidth]{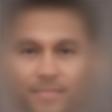}\hfill
		\includegraphics[width=0.062\linewidth]{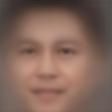}\hfill
		\includegraphics[width=0.062\linewidth]{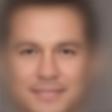}\hfill
		\includegraphics[width=0.062\linewidth]{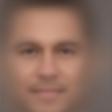}\hfill
		\includegraphics[width=0.062\linewidth]{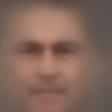}
		\end{minipage}
	    \end{subfigure}\\
	    
	    \begin{subfigure}[b]{\linewidth}
		\centering
		\begin{minipage}[c]{0.08\textwidth}
		\tiny{DebFace-R}
	    \end{minipage}\hfill
	    \begin{minipage}[c]{0.92\textwidth}
		\includegraphics[width=0.062\linewidth]{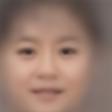}\hfill
		\includegraphics[width=0.062\linewidth]{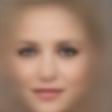}\hfill
		\includegraphics[width=0.062\linewidth]{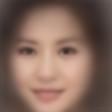}\hfill
		\includegraphics[width=0.062\linewidth]{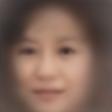}\hfill
		\includegraphics[width=0.062\linewidth]{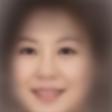}\hfill
		\includegraphics[width=0.062\linewidth]{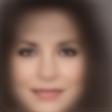}\hfill
		\includegraphics[width=0.062\linewidth]{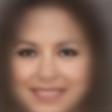}\hfill
		\includegraphics[width=0.062\linewidth]{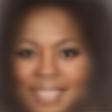}\hfill
		\includegraphics[width=0.062\linewidth]{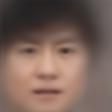}\hfill
		\includegraphics[width=0.062\linewidth]{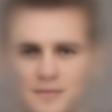}\hfill
		\includegraphics[width=0.062\linewidth]{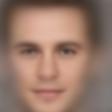}\hfill
		\includegraphics[width=0.062\linewidth]{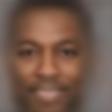}\hfill
		\includegraphics[width=0.062\linewidth]{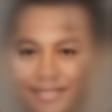}\hfill
		\includegraphics[width=0.062\linewidth]{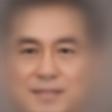}\hfill
		\includegraphics[width=0.062\linewidth]{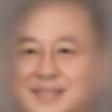}\hfill
		\includegraphics[width=0.062\linewidth]{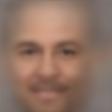}
		\end{minipage}
	    \end{subfigure}\\
	    
	    \begin{subfigure}[b]{\linewidth}
		\centering
		\begin{minipage}[c]{0.08\textwidth}
		\tiny{DebFace-A}
	    \end{minipage}\hfill
	    \begin{minipage}[c]{0.92\textwidth}
		\includegraphics[width=0.062\linewidth]{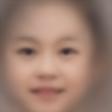}\hfill
		\includegraphics[width=0.062\linewidth]{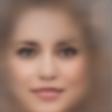}\hfill
		\includegraphics[width=0.062\linewidth]{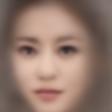}\hfill
		\includegraphics[width=0.062\linewidth]{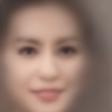}\hfill
		\includegraphics[width=0.062\linewidth]{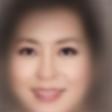}\hfill
		\includegraphics[width=0.062\linewidth]{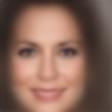}\hfill
		\includegraphics[width=0.062\linewidth]{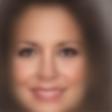}\hfill
		\includegraphics[width=0.062\linewidth]{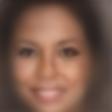}\hfill
		\includegraphics[width=0.062\linewidth]{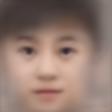}\hfill
		\includegraphics[width=0.062\linewidth]{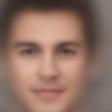}\hfill
		\includegraphics[width=0.062\linewidth]{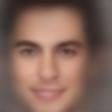}\hfill
		\includegraphics[width=0.062\linewidth]{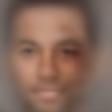}\hfill
		\includegraphics[width=0.062\linewidth]{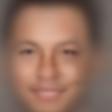}\hfill
		\includegraphics[width=0.062\linewidth]{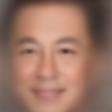}\hfill
		\includegraphics[width=0.062\linewidth]{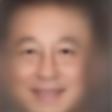}\hfill
		\includegraphics[width=0.062\linewidth]{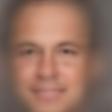}
		\end{minipage}
	    \end{subfigure}\\
	    \caption{\footnotesize Reconstructed Images using Face and Demographic Representations. The first row is the original face images. From the second row to the bottom, the face images are reconstructed from 2) BaseFace; 3) DebFace-ID; 4) DebFace-G; 5) DebFace-R; 6) DebFace-A. Zoom in for details.}
	    \label{img:reconstruct}
\end{figure}

\subsection{Analysis of Disentanglement}
We notice that DebFace still suffers unequal performance in different demographic groups. It is because there are other latent variables besides the demographics, such as image quality or capture conditions that could lead to biased performance. Such variables are difficult to control in pre-collected large face datasets. In the framework of DebFace, it is also related to the degree of feature disentanglement. A fully disentangling is supposed to completely remove the factors of bias from demographic information. 
To illustrate the feature disentanglement of DebFace, we show the demographic discriminative ability of face representations by using these features to estimate gender, age, and race. Specifically, we first extract identity features of images from the testing set in Sec.~\ref{sec:datasets} and split them into training and testing sets. Given demographic labels, the face features are fed into a two-layer fully-connected network, learning to classify one of the demographic attributes. 
Tab.~\ref{tab:demog_classify} reports the demographic classification accuracy on the testing set. For all three demographic estimations, DebFace-ID presents much lower accuracies than BaseFace, indicating the decline of demographic information in DebFace-ID.
We also plot the distribution of identity representations in the feature space of BaseFace and DebFace-ID. From the testing set in Sec.~\ref{sec:face_verify}, we randomly select 50 subjects in each demographic group and one image of each subject. BaseFace and DebFace-ID are extracted from the selected image set and are then projected from $512$-\textit{dim} to $2$-\textit{dim} by T-SNE. Fig.~\ref{fig:tsne_face} shows their T-SNE feature distributions. We observe that BaseFace presents clear demographic clusters, while the demographic clusters of DebFace-ID, as a result of disentanglement, mostly overlap with each other.

To visualize the disentangled feature representations of DebFace, we train a decoder that reconstructs face images from the representations. Four face decoders are trained separately for each disentangled component, {\it i.e.}, gender, age, race, and ID. In addition, we train another decoder to reconstruct faces from BaseFace for comparison. As shown in Fig.~\ref{img:reconstruct}, both BaseFace and DebFace-ID maintain the identify features of the original faces, while DebFace-ID presents less demographic characteristics. No race or age, but gender features can be observed on faces reconstructed from DebFace-G. Meanwhile, we can still recognize race and age attributes on faces generated from DebFace-R and DebFace-A.

\begin{figure}[!t]
\centerline{\begin{minipage}[t!]{\textwidth}
  \begin{minipage}[b]{0.44\textwidth}
    \centering
    \captionof{table}{\footnotesize Demographic Classification Accuracy (\%) by face features.}
    \label{tab:demog_classify}
    \scalebox{0.9}{
    \begin{tabular}{c ccc}
        \toprule
        Method & Gender & Race & Age \\
        \midrule
        BaseFace & 95.27 & 89.82 & 78.14\\
        DebFace-ID &  73.36 & 61.79 & 49.91\\
        \bottomrule
    \end{tabular}}
  \end{minipage}
  \hfill
  \begin{minipage}[b]{0.5\textwidth}
    \centering
    \captionof{table}{\footnotesize Face Verification Accuracy (\%) on RFW dataset.}
    \label{tab:rfw}
    \scalebox{0.85}{
    \begin{tabular}{c cccc c}
        \toprule
        Method & White & Black & Asian & Indian & Biasness \\
        \midrule
        \cite{wang2020mitigating} & 96.27 & 94.68 & 94.82 & 95.00 & 0.93 \\
        DebFace-ID & 95.95 & 93.67 & 94.33 & 94.78 & 0.83\\
        \bottomrule
      \end{tabular}}
    \end{minipage}
  \end{minipage}}
\end{figure}  

\begin{table}[t]
    \centering
    \caption{\footnotesize{Verification Performance on LFW, IJB-A, and IJB-C.}}
    \label{tab:lfw_ijba_ijbc}
    \scalebox{0.8}{
    \begin{tabularx}{\linewidth}{X c | X c c c c}
        \toprule
        \multirow{2}{*}{Method} & \multirow{2}{*}{LFW (\%)} & \multirow{2}{*}{Method} & IJB-A (\%) & \multicolumn{3}{c}{IJB-C @ FAR (\%)}\\
        \cline{5-7}
        & & & 0.1\% FAR & 0.001\% & 0.01\% & 0.1\% \\
        \midrule  
        DeepFace+~\cite{taigman2014deepface} & $97.35$ & Yin~\etal~\cite{yin2017multi} & $73.9 \pm 4.2$ & - & - & 69.3 \\
        CosFace~\cite{wang2018cosface} & $99.73$ & Cao~\etal~\cite{cao2018vggface2} & $90.4 \pm 1.4$ & $74.7$ & $84.0$ & $91.0$ \\
        ArcFace~\cite{Deng_2019_CVPR} & $99.83$ & Multicolumn~\cite{xie2018multicolumn} & $92.0 \pm 1.3$ & $77.1$ & $86.2$ & $92.7$ \\
        PFE~\cite{shi2019probabilistic} & $99.82$ & PFE~\cite{shi2019probabilistic} & $95.3 \pm 0.9$ & $89.6$ & $93.3$ & $95.5$ \\
        \midrule
        \textit{BaseFace} & $99.38$ & \textit{BaseFace} & $90.2 \pm 1.1$ & $80.2$ & $88.0$ & $92.9$ \\
        \textit{DebFace-ID} & $98.97$ & \textit{DebFace-ID} & $87.6 \pm 0.9$ & $82.0$ & $88.1$ & $89.5$ \\
        \textit{DemoID} & $99.50$ & \textit{DemoID} & $92.2 \pm 0.8$ & $83.2$ & $89.4$ & $92.9$ \\
        \bottomrule
    \end{tabularx}}
\end{table}

\subsection{Face Verification on Public Testing Datasets}
We report the performance of three different settings, using 1) BaseFace, the same baseline in Sec.~\ref{sec:face_verify}, 2) DebFace-ID, and 3) the fused representation DemoID.
Table~\ref{tab:lfw_ijba_ijbc} reports face verification results on on three public benchmarks: LFW, IJB-A, and IJB-C. 
On LFW, DemoID outperforms BaseFace while maintaining similar accuracy compared to SOTA algorithms. 
On IJB-A/C, DemoID outperforms all prior works except PFE~\cite{shi2019probabilistic}. 
Although DebFace-ID shows lower discrimination, TAR at lower FAR on IJB-C is higher than that of BaseFace.
To evaluate DebFace on a racially balanced testing dataset RFW~\cite{Wang_2019_ICCV} and compare with the work~\cite{wang2020mitigating}, we train a DebFace model on BUPT-Balancedface~\cite{wang2020mitigating} dataset. The new model is trained to reduce racial bias by disentangling ID and race. Tab.~\ref{tab:rfw} reports the verification results on RFW. 
While DebFace-ID gives a slightly lower face verification accuracy, it improves the biasness over~\cite{wang2020mitigating}. 

\begin{figure} [t!]
	    \captionsetup{font=footnotesize}
	    \centering
		\begin{subfigure}[b]{0.23\linewidth}
		\centering
		\includegraphics[width=\linewidth]{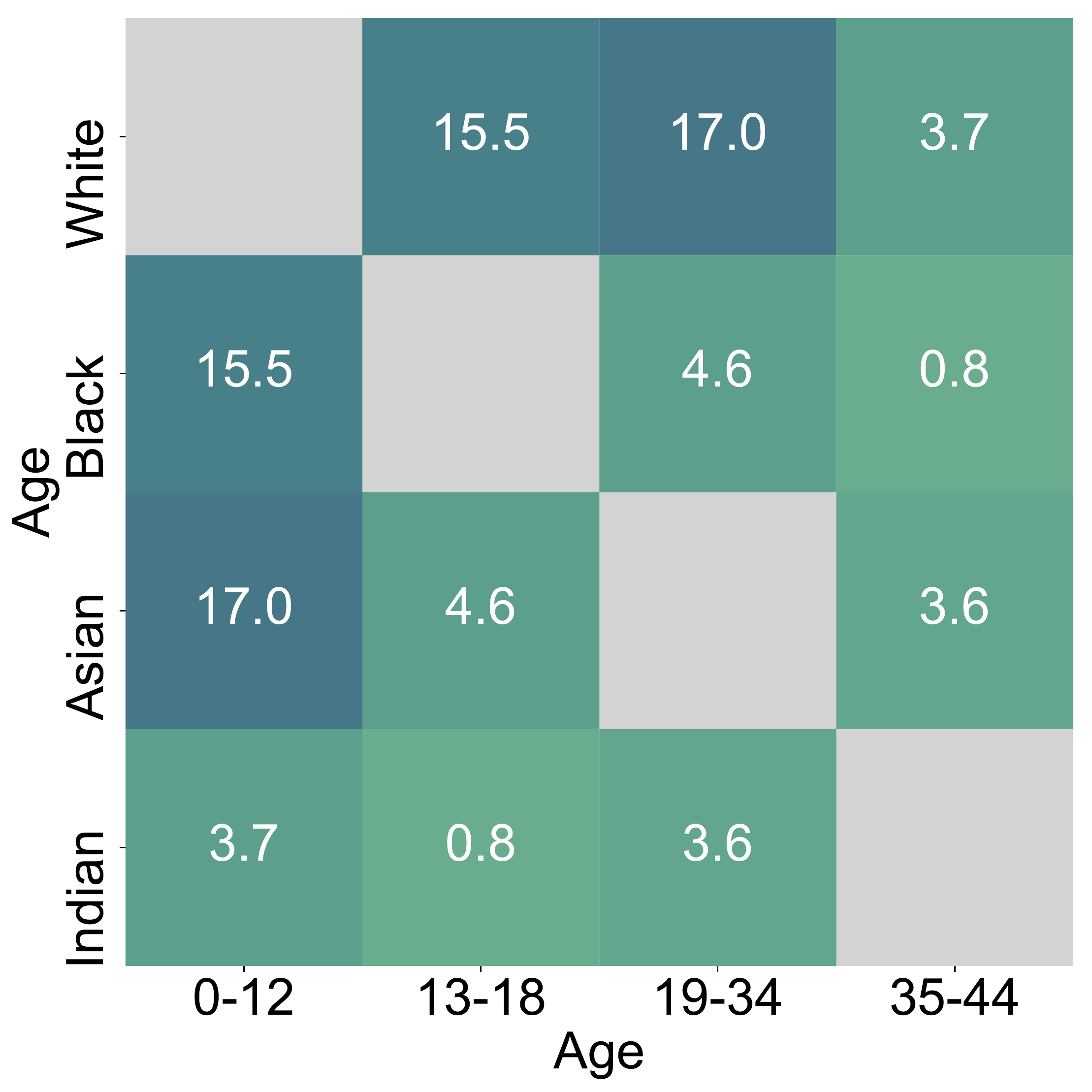}
	    \caption{{\scriptsize BaseFace: Race}}
	    \label{fig:cross_race_base}
	    \end{subfigure}\hfill    
	    \begin{subfigure}[b]{0.23\linewidth}
	    \centering
	    \includegraphics[width=\linewidth]{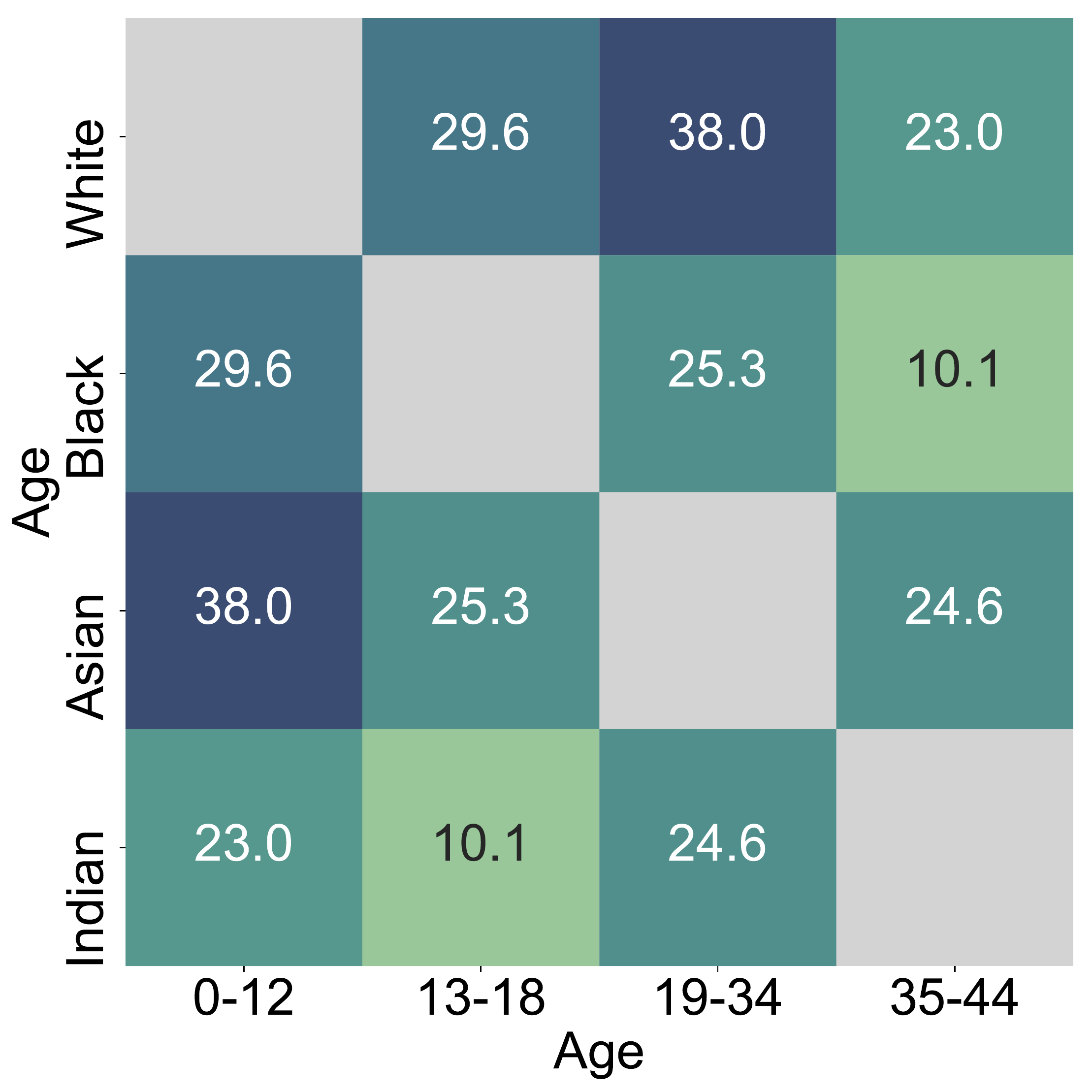}
	    \caption{{\scriptsize DebFace-ID: Race}}
	    \label{fig:face_race_deb}
	    \end{subfigure}\hfill
	    \begin{subfigure}[b]{0.25\linewidth}
		\centering
		\includegraphics[width=\linewidth]{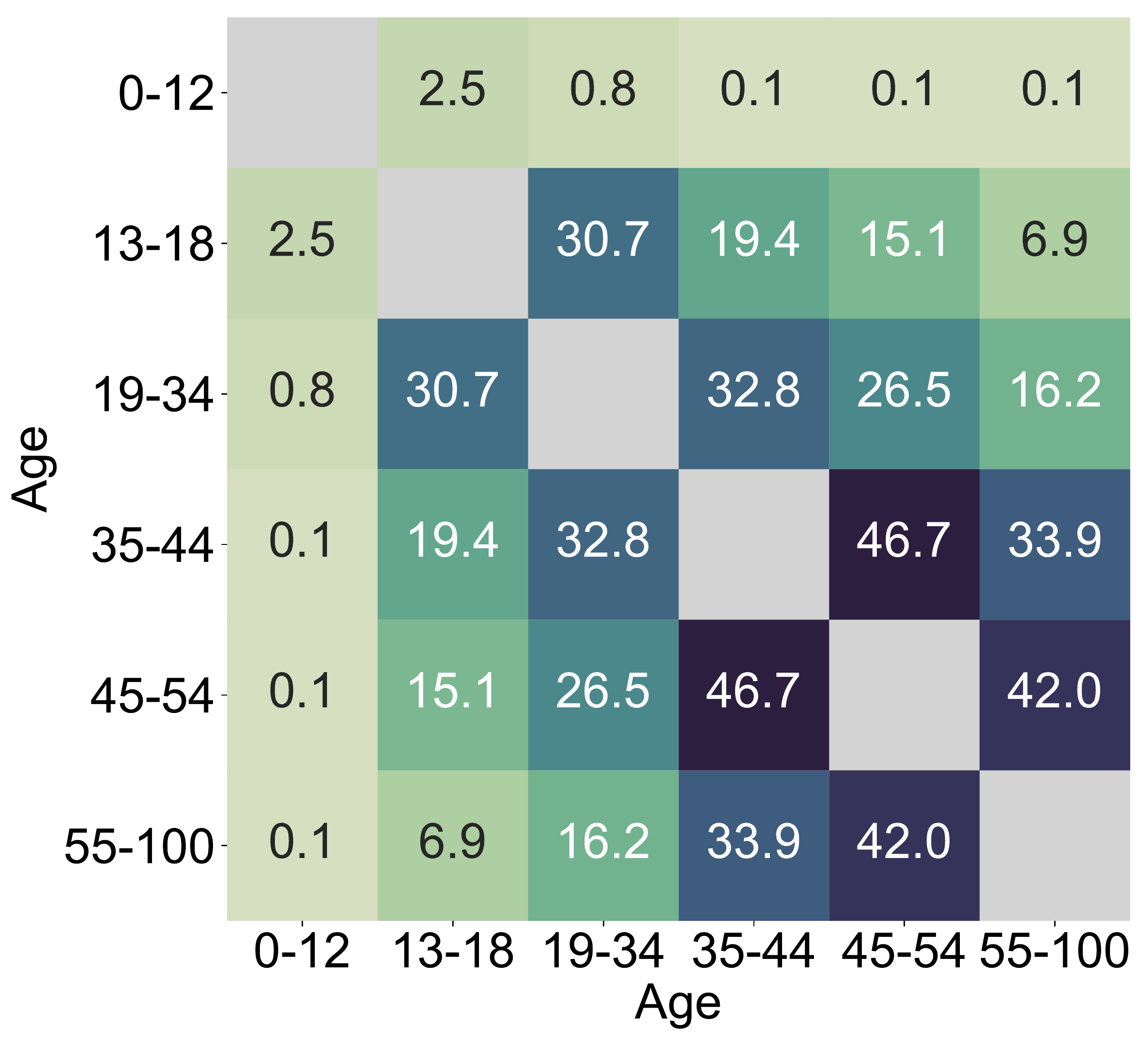}
	    \caption{{\scriptsize BaseFace: Age}}
	    \label{fig:cross_age_base}
	    \end{subfigure}\hfill    
	    \begin{subfigure}[b]{0.25\linewidth}
	    \centering
	    \includegraphics[width=\linewidth]{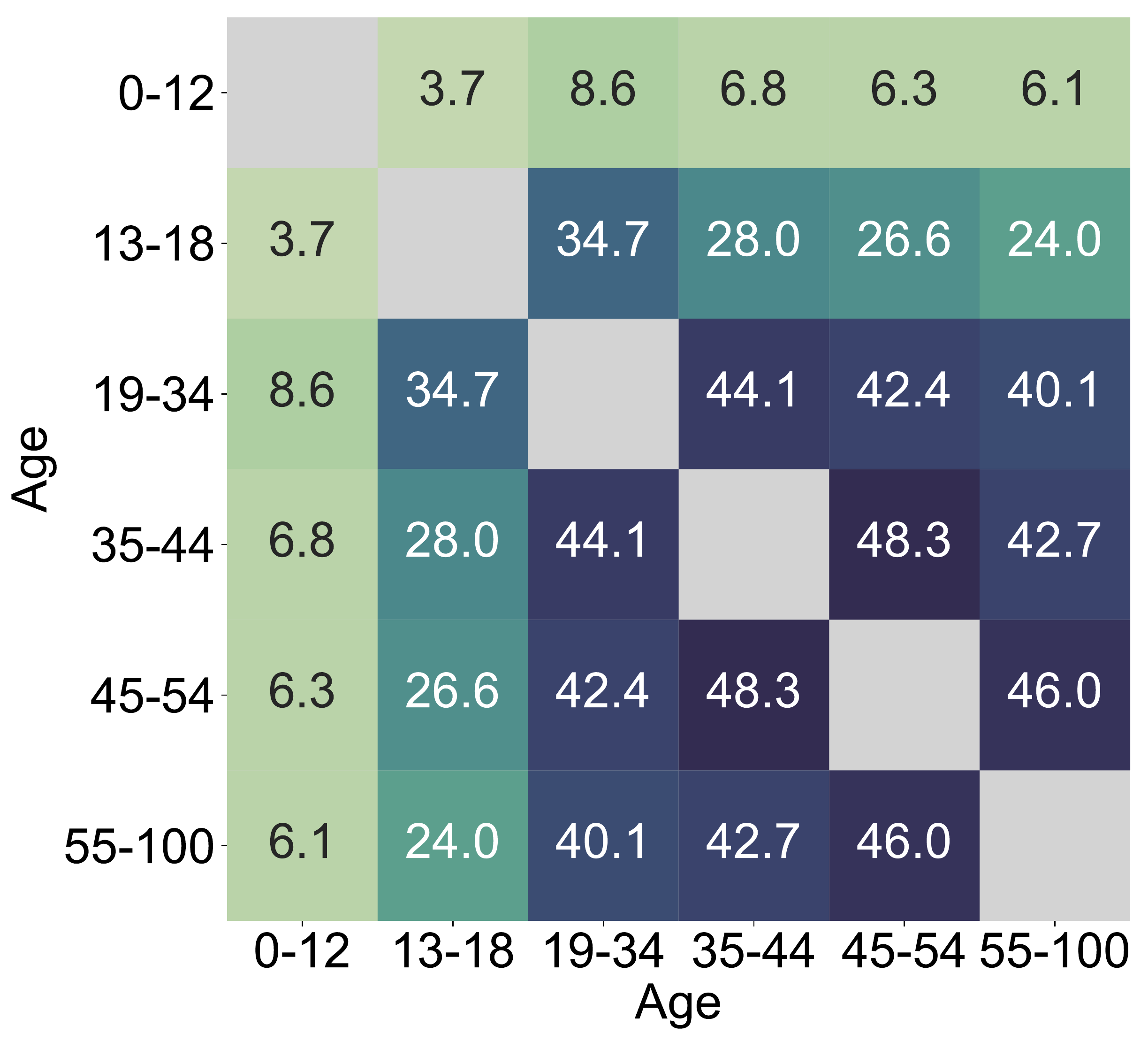}
	    \caption{{\scriptsize DebFace-ID: Age}}
	    \label{fig:face_age_deb}
	    \end{subfigure}\\
	    \caption{\footnotesize{The percentage of false accepted cross race or age pairs at 1\% FAR.}}
	    \label{fig:cross_race_age}
\end{figure}

We observe that DebFace-ID is less discriminative than BaseFace, or DemoID, since demographics are essential components of face features.
To understand the deterioration of DebFace, we analyse the effect of demographic heterogeneity on face verification by showing the tendency for one demographic group to experience a false accept error relative to another group. For any two demographic cohorts, we check the number of falsely accepted pairs that are from different groups at $1\%$ FAR. Fig.~\ref{fig:cross_race_age} shows the percentage of such falsely accepted demographic-heterogeneous pairs. Compared to BaseFace, DebFace exhibits more cross-demographic pairs that are falsely accepted, resulting in the performance decline on demographically biased datasets. Due to the demographic information reduction, DebFace-ID is more susceptible to errors between demographic groups.
In the sense of de-biasing, it is preferable to decouple demographic information from identity features. 
However, if we prefer to maintain the overall performance across all demographics, we can still aggregate all the relevant information.
It is an application-dependent trade-off between accuracy and de-biasing.
DebFace balances the accuracy vs. bias trade-off by generating both debiased identity and debiased demographic representations, which may be aggregated into DemoID if bias is less of a concern.

\section{Conclusion}
We present a de-biasing face recognition network (DebFace) to mitigate demographic bias in face recognition. 
DebFace adversarially learns the disentangled representation for gender, race, and age estimation, and face recognition simultaneously. 
We empirically demonstrate that DebFace can not only reduce bias in face recognition but in demographic attribute estimation as well. 
Our future work will explore an aggregation scheme to combine race, gender, and identity without introducing algorithmic and dataset bias.

\noindent\textbf{Acknowledgement:} This work is supported by U.S.~Department of Commerce (\#60NANB19D154), National Institute of Standards and Technology. The authors thank reviewers, area chairs, Dr.~John J.~Howard, and Dr.~Yevgeniy Sirotin for offering constructive comments.

\clearpage
%
%
{\small
\bibliographystyle{splncs04}
\bibliography{egbib}
}

\end{document}


\pagestyle{headings}
\mainmatter
\def\ECCVSubNumber{6719}  

\title{Jointly De-biasing Face Recognition and Demographic Attribute Estimation  \\ (Supplementary Material)} 

\titlerunning{DebFace}
%
%
\author{Sixue Gong\quad\quad Xiaoming Liu\quad\quad Anil K. Jain\\
{\tt\small \{gongsixu, liuxm, jain\}@msu.edu}
}
\authorrunning{Sixue Gong et al.}
%
\institute{Michigan State University}
\maketitle

\thispagestyle{empty}
In this supplementary material we include: (1) Section \ref{sec:datasets}: the statistics of datasets used in the experiments, (2) Section \ref{sec:demog}: implementation details and performance of the three demographic models trained to label MS-Celeb-1M, (3) Section \ref{sec:scores}: distributions of the scores of the imposter pairs across homogeneous versus heterogeneous, (4) Section \ref{sec:crossage}: performance comparisons of cross-age face recognition.

\section{Datasets}

Table~\ref{tab:datasets} reports the statistics of training and testing datasets involved in the experiments, including the total number of face images, the total number of subjects (identities), and whether the dataset contains the annotation of gender, age, race, or identity (ID).

\label{sec:datasets}
\begin{table}[t]
    \centering
    \captionsetup{font=footnotesize}
    \caption{Statistics of training and testing datasets used in the paper.}
    \label{tab:datasets}
    \begin{threeparttable}
    \scalebox{0.9}{
    \begin{tabularx}{1.05\linewidth}{c c c c c c c c c c}
        \toprule
        \multirow{2}{*}{Dataset} && \multirow{2}{*}{\# of Images} && \multirow{2}{*}{\# of Subjects} && \multicolumn{4}{c}{Contains the label of}\\
        \cline{7-10}
        && && && Gender & Age & Race & ID \\
        \midrule  
        CACD~\cite{chen2014cross} && $163,446$ && $2,000$ && No & Yes & No & Yes \\
        IMDB~\cite{rothe2018deep} && $460,723$ && $20,284$ && Yes & Yes & No & Yes \\
        UTKFace~\cite{zhifei2017cvpr} && $24,106$ && - && Yes & Yes & Yes & No \\
        AgeDB~\cite{moschoglou2017agedb} && $16,488$ && $567$ && Yes & Yes & No & Yes \\
        AFAD~\cite{niu2016ordinal} && $165,515$ && - && Yes & Yes & Yes\tnote{$a$} & No \\
        AAF~\cite{cheng2019exploiting} && $13,322$ && $13,322$ && Yes & Yes & No & Yes \\
        FG-NET~\footnote{\url{https://yanweifu.github.io/FG\_NET\_data}} && $1,002$ && $82$ && No & Yes & No & Yes \\
        RFW~\cite{Wang_2019_ICCV} && $665,807$ && - && No & No & Yes & Partial \\
        IMFDB-CVIT~\cite{imfdb} && $34,512$ && $100$ && Yes & Age Groups & Yes\tnote{*} & Yes \\
        Asian-DeepGlint~\cite{AsianCeleb} && $2,830,146$ && $93,979$ && No & No & Yes\tnote{$a$} & Yes \\
        MS-Celeb-1M~\cite{guo2016ms} && $5,822,653$ && $85,742$ && No & No & No & Yes \\
        PCSO~\cite{deb2017face} && $1,447,607$ && $5,749$ && Yes & Yes & Yes & Yes \\
        LFW~\cite{huang2008labeled} && $13,233$ && $5,749$ && No & No & No & Yes \\
        IJB-A~\cite{klare2015pushing} && $25,813$ && $500$ && Yes & Yes & Skin Tone & Yes \\
        IJB-C~\cite{maze2018iarpa} && $31,334$ && $3,531$ && Yes & Yes & Skin Tone & Yes \\
        \bottomrule
    \end{tabularx}}
    
    \begin{tablenotes}\scriptsize
	\item[$a$] East Asian
    \item[*] Indian
    \end{tablenotes}
    \end{threeparttable}
\end{table}

\section{Demographic Estimation}
\label{sec:demog}
We train three demographic estimation models to annotate age, gender, and race information of the face images in MS-Celeb-1M for training DebFace. For all three models, we randomly sample equal number of images from each class and set the batch size to $300$. The training process finishes at $35K^{th}$ iteration. All hyper-parameters are chosen by testing on a separate validation set. Below gives the details of model learning and estimation performance of each demographic.

\begin{figure}[t!]
	    \captionsetup{font=footnotesize}
	    \centering
	    \begin{subfigure}[b]{0.33\linewidth}
	    \includegraphics[width=\linewidth]{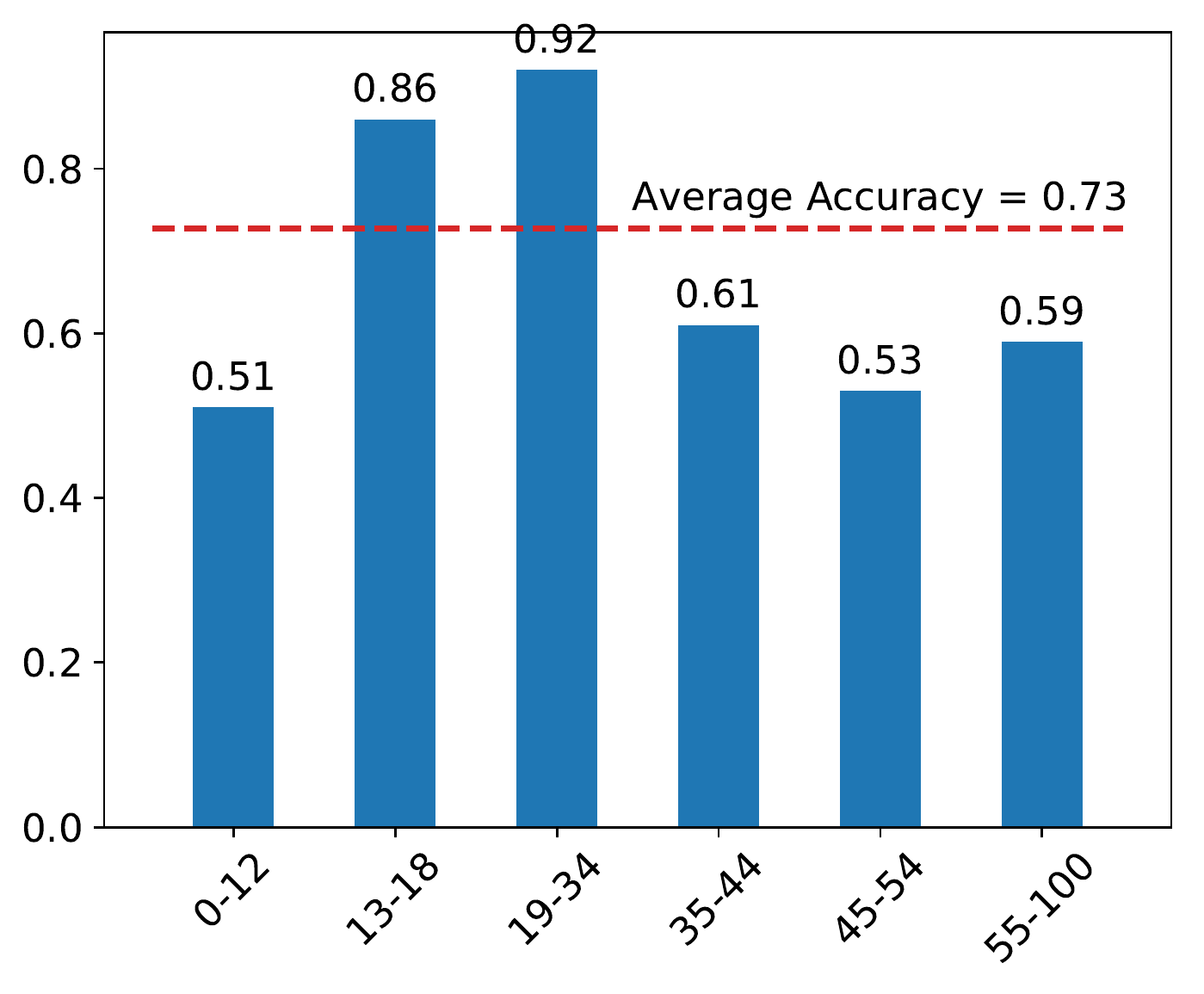}
	    \caption{{\footnotesize Age}}
	    \label{fig:age_acc}
	    \end{subfigure}\hfill		
		\begin{subfigure}[b]{0.33\linewidth}
		\includegraphics[width=\linewidth]{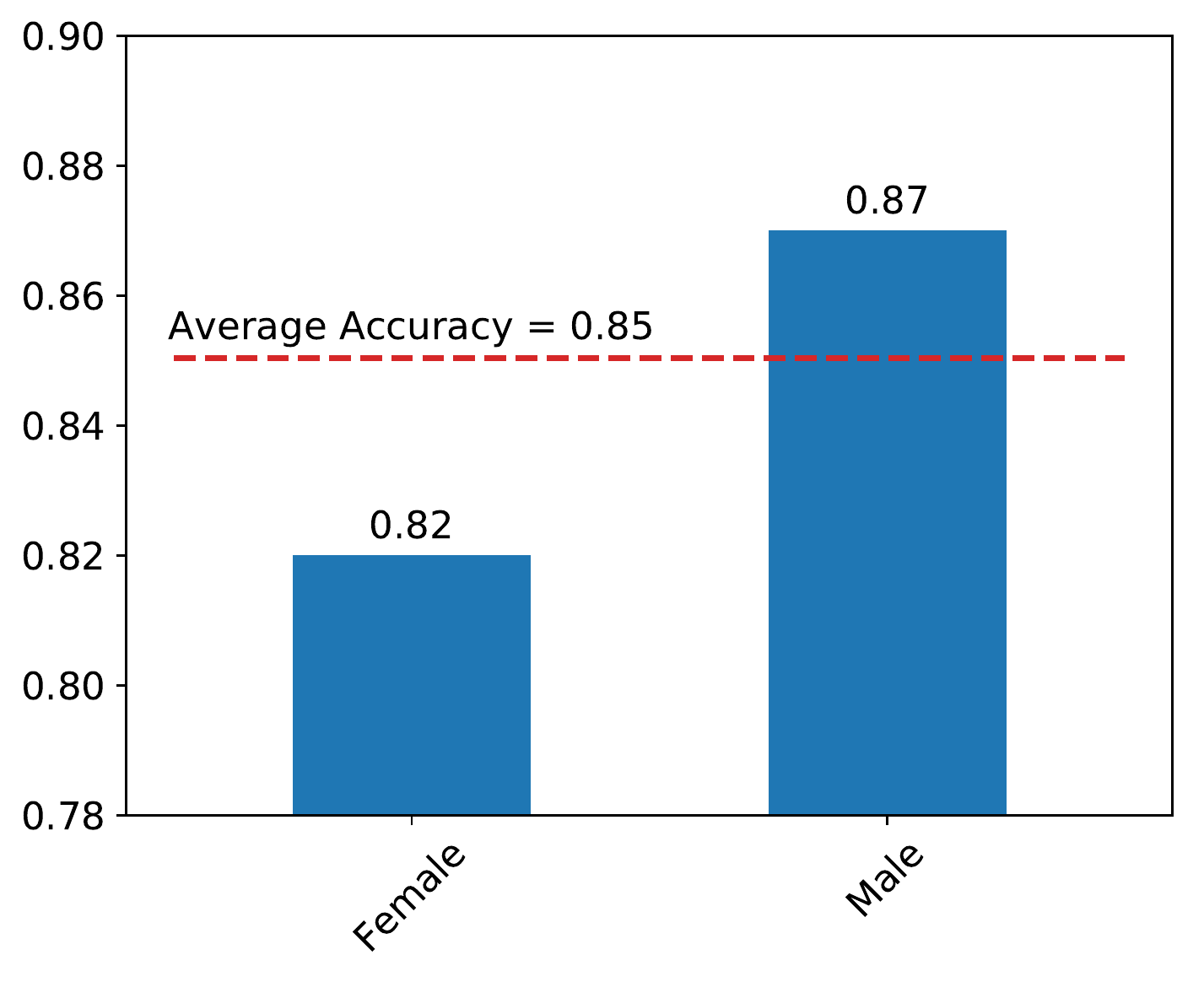}
	    \caption{{\footnotesize Gender}}
	    \label{fig:gender_acc}
	    \end{subfigure}\hfill    
	    \begin{subfigure}[b]{0.33\linewidth}
	    \includegraphics[width=\linewidth]{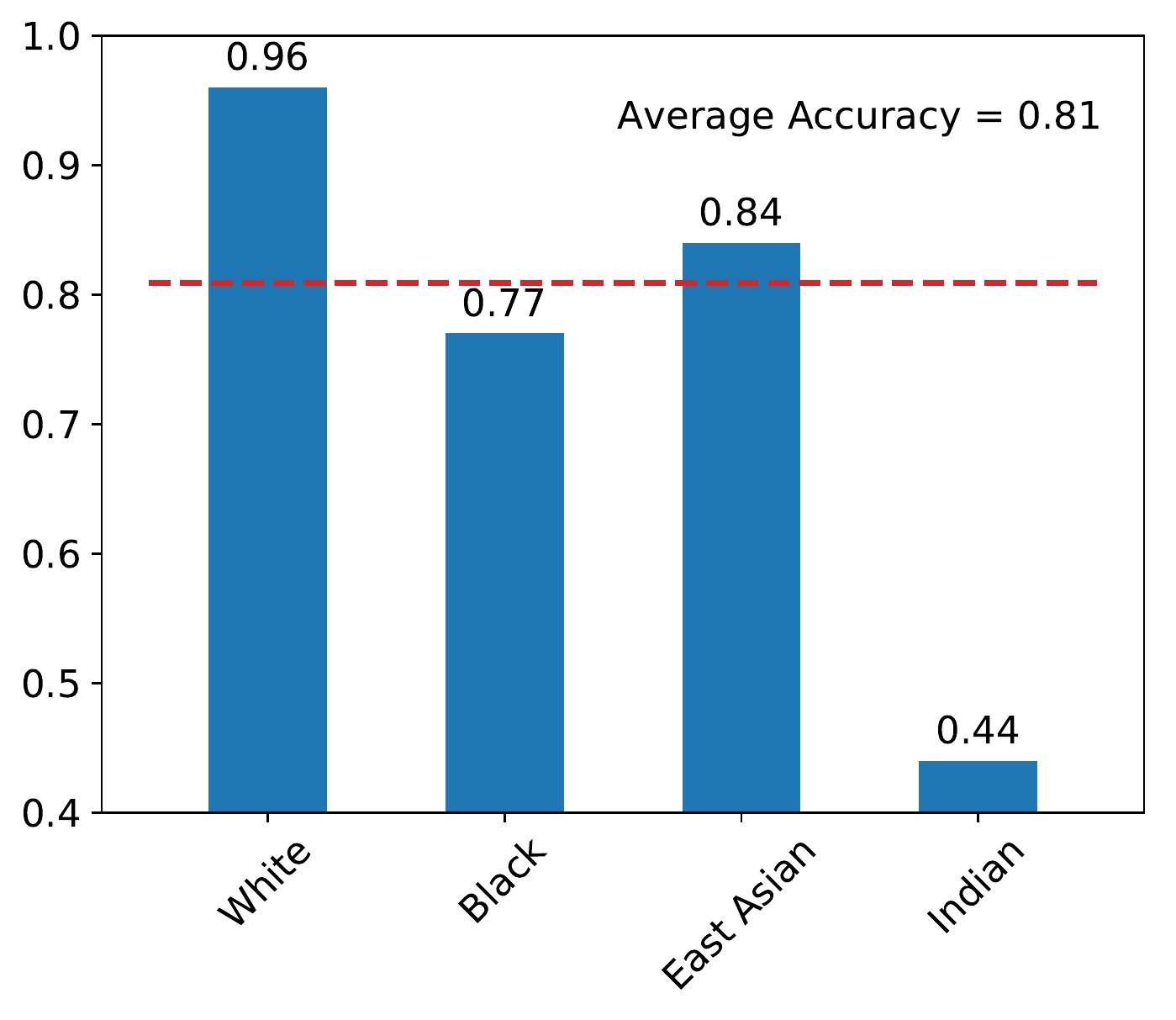}
	    \caption{{\footnotesize Race}}
	    \label{fig:race_acc}
	    \end{subfigure}\\	    
	    \caption{\footnotesize Demographic Attribute Classification Accuracy on each group. The red dashed line refers to the average accuracy on all images in the testing set.}
\end{figure}

\textbf{Gender:} We combine IMDB, UTKFace, AgeDB, AFAD, and AAF datasets for learning the gender estimation model. Similar to age, $90\%$ of the images in the combined datasets are used for training, and the remaining $10\%$ are used for validation. Table~\ref{tab:gender_stats} reports the total number of female and male face images in the training and testing set. More images belong to male faces in both training and testing set. Figure~\ref{fig:gender_acc} shows the gender estimation performance on the validation set. The performance on male images is slightly better than that on female images.

\vspace{7px}
\centerline{\begin{minipage}[t!]{\linewidth}
  \begin{minipage}[b]{0.39\linewidth}
    \centering
    \captionof{table}{\footnotesize Gender distribution of the datasets for gender estimation.}
    \label{tab:gender_stats}
    \scalebox{0.9}{
    \begin{tabular}{P{1.2cm}P{0.2cm} P{1.2cm} P{1.2cm} P{1.2cm}}
        \toprule
        \multirow{2}{*}{Dataset} && \multicolumn{2}{c}{\# of Images}\\
        \cline{3-4}
        && Male & Female \\
        \midrule  
        Training && 321,590 & 229,000 \\
        Testing && 15,715 & 10,835 \\
        \bottomrule
    \end{tabular}}
  \end{minipage}
  \hfill
  \begin{minipage}[b]{0.53\linewidth}
    \centering
    \captionof{table}{\footnotesize Race distribution of the datasets for race estimation.}
    \label{tab:race_stats}
    \scalebox{0.9}{
    \begin{tabular}{c c c c c c}
        \toprule
        \multirow{2}{*}{Dataset} && \multicolumn{4}{c}{\# of Images}\\
        \cline{3-6}
        && White & Black & East Asian & Indian \\
        \midrule  
        Training && 468,139 & 150,585 & 162,075 & 78,260 \\
        Testing && 9,469 & 4,115 & 3,336 & 3,748 \\
        \bottomrule
    \end{tabular}}
    \end{minipage}
  \end{minipage}}
  \vspace{14px}

\begin{table}[t]
    \centering
    \caption{Age distribution of the datasets for age estimation}
    \label{tab:age_stats}
    \scalebox{1.0}{
    \begin{tabular}{P{1.2cm}P{0.2cm} P{1.2cm} P{1.2cm} P{1.2cm} P{1.2cm} P{1.2cm} P{1.2cm} P{1.2cm}}
        \toprule
        \multirow{2}{*}{Dataset} && \multicolumn{6}{c}{\# of Images in the Age Group}\\
        \cline{3-8}
        && 0-12 & 13-18 & 19-34 & 35-44 & 45-54 & 55-100 \\
        \midrule  
        Training && 9,539 & 29,135 & 353,901 & 171,328 & 93,506 & 59,599 \\
        Testing && 1,085 & 2,681 & 13,848 & 8,414 & 5,479 & 4,690 \\  
        \bottomrule
    \end{tabular}}
\end{table}

\begin{figure}
\captionsetup{font=footnotesize}
    \centering
	    \begin{subfigure}[b]{0.49\linewidth}
	    \includegraphics[width=\linewidth]{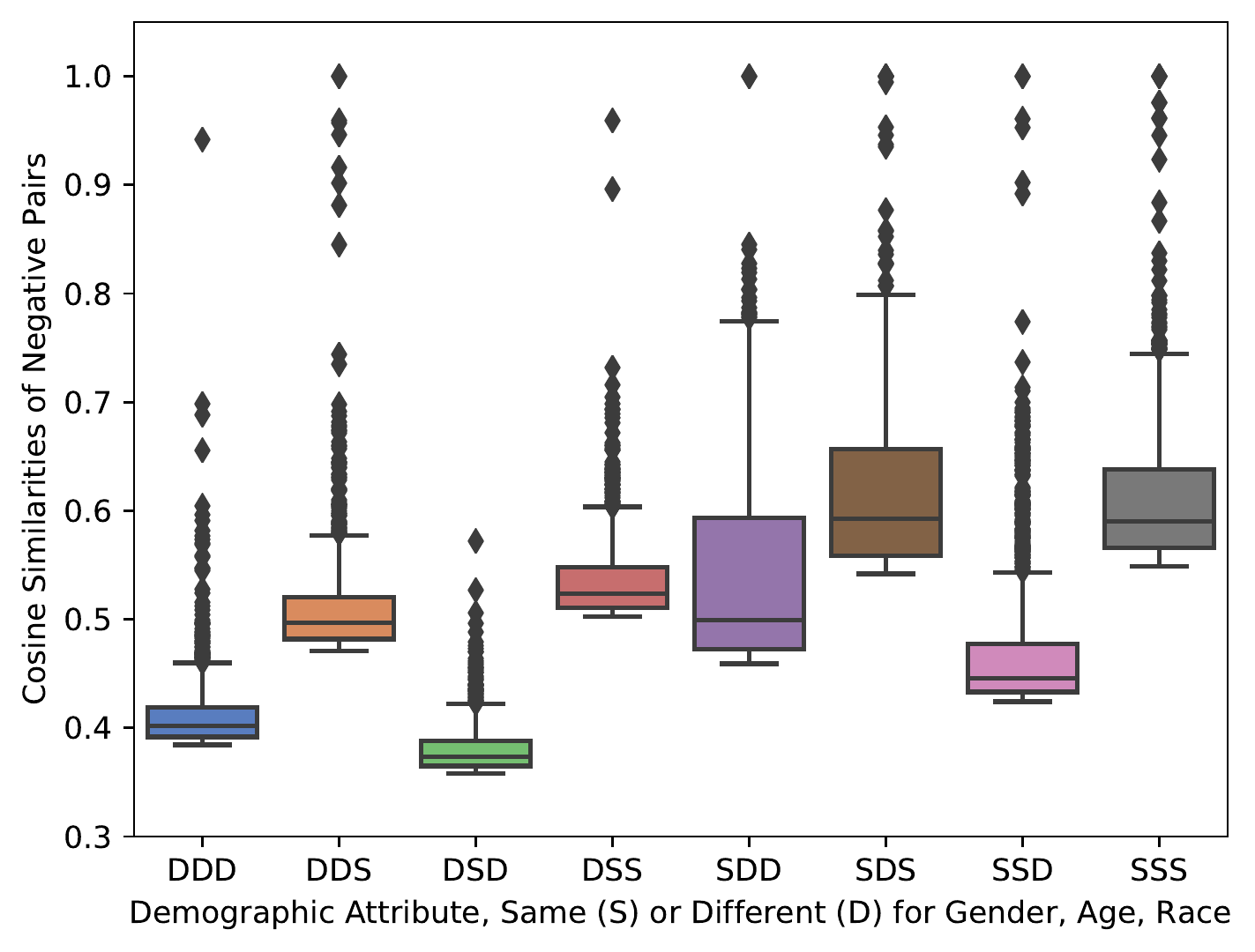}
	    \caption{{\footnotesize BaseFace}}
	    \label{fig:base_score}
	    \end{subfigure}\hfill
	    \begin{subfigure}[b]{0.49\linewidth}
	    \includegraphics[width=\linewidth]{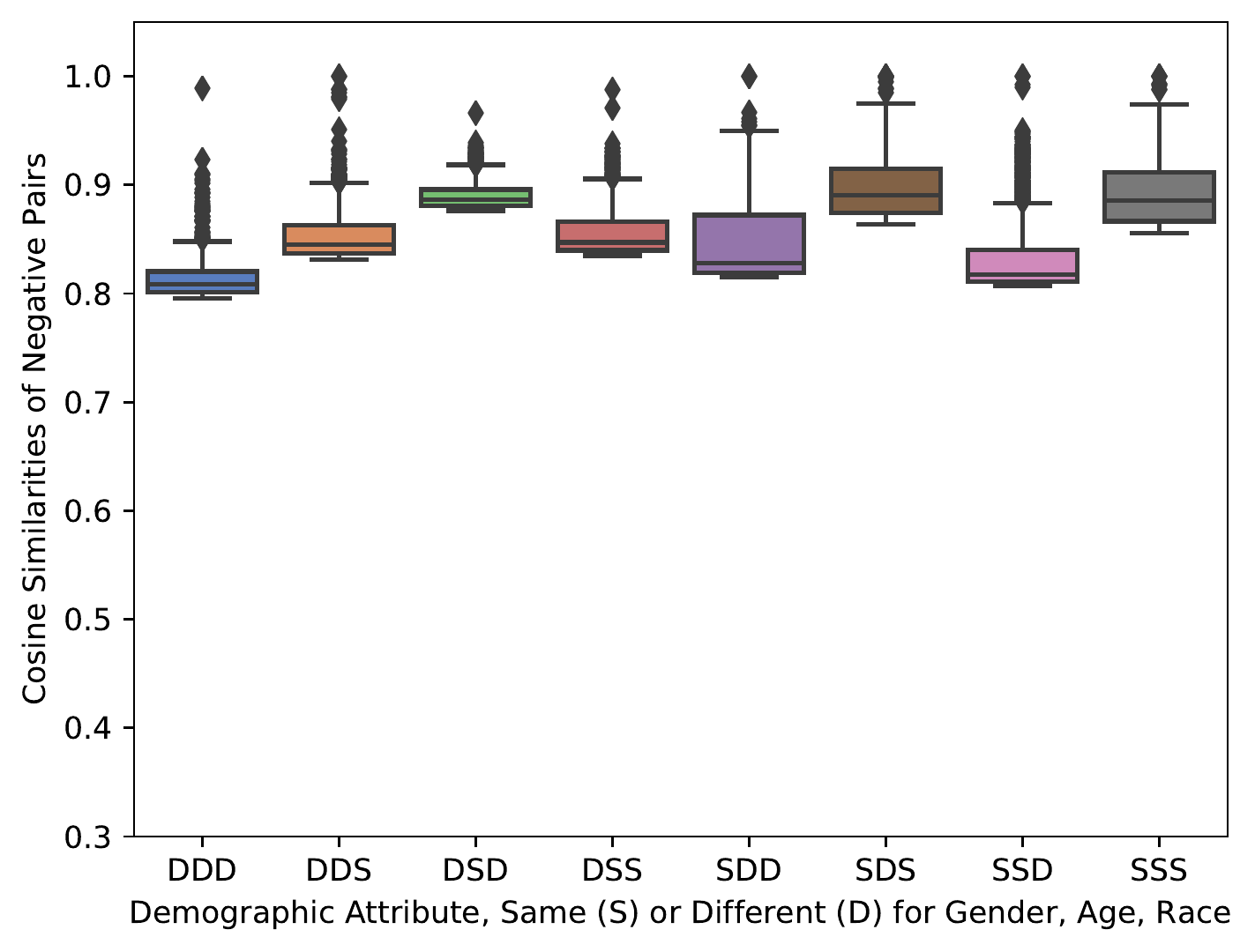}
	    \caption{{\footnotesize DebFace}}
	    \label{fig:deb_score}
	    \end{subfigure}\\
	\caption{\footnotesize BaseFace and DebFace distributions of the similarity scores of the imposter pairs across homogeneous versus heterogeneous gender, age, and race categories.}
\end{figure}

\textbf{Race:} We combine AFAD, RFW, IMFDB-CVIT, and PCSO datasets for training the race estimation model. UTKFace is used as validation set. Table~\ref{tab:race_stats} reports the total number of images in each race category of the training and testing set. Similar to age and gender, the performance of race estimation is highly correlated to the race distribution in the training set. Most of the images are within the White group, while the Indian group has the least number of images. Therefore, the performance on White faces is much higher than that on Indian faces.

\textbf{Age:} We combine CACD, IMDB, UTKFace, AgeDB, AFAD, and AAF datasets for learning the age estimation model. $90\%$ of the images in the combined datasets are used for training, and the remaining $10\%$ are used for validation. Table~\ref{tab:age_stats} reports the total number of images in each age group of the training and testing set, respectively. Figure~\ref{fig:age_acc} shows the age estimation performance on the validation set. The majority of the images come from the age 19 to 34 group. Therefore, the age estimation performs the best on this group. The performance on the young children and middle to old age group is significantly worse than the majority group.

It is clear that all the demographic models present biased performance with respect to different cohorts. These demographic models are used to label the MS-Celeb-1M for training DebFace. Thus, in addition to the bias from the dataset itself, we also 
add label bias to it. Since DebFace employs supervised feature disentanglement, we only strive to reduce the data bias instead of the label bias. 

\section{Distributions of Scores}
\label{sec:scores}

We follow the work of~\cite{howard2019effect} that investigates the effect of demographic homogeneity and heterogeneity on face recognition. We first randomly select images from CACD, AgeDB, CVIT, and Asian-DeepGlint datasets, and extract the corresponding feature vectors by using the models of BaseFace and DebFace, respectively. Given their demographic attributes, we put those images into separate groups depending on whether their gender, age, and race are the same or different. For each group, a fixed false alarm rate (the percentage of the face pairs from the same subjects being falsely verified as from different subjects) is set to $1\%$.
Among the falsely verified pairs, we plot the top $10^{th}$ percentile scores of the negative face pairs (a pair of face images that are from different subjects) given their demographic attributes. As shown in Fig.~\ref{fig:base_score} and Fig.~\ref{fig:deb_score}, we observe that the similarities of DebFace are higher than those of BaseFace. One of the possible reasons is that the demographic information is disentangled from the identity features of DebFace, increasing the overall pair-wise similarities between faces of different identities. In terms of de-biasing, DebFace also reflects smaller differences of the score distribution with respect to the homogeneity and heterogeneity of demographics.

\section{Cross-age Face Recognition}
\label{sec:crossage}

\begin{table}[t]
    \centering
    \caption{Evaluation Results (\%) of Cross-Age Face Recognition}
    \label{tab:crossage}
    \scalebox{1.0}{
    \begin{tabular}{P{1.5cm}P{0.2cm} P{1.5cm} P{1.5cm}}
        \toprule
        \multirow{2}{*}{Method} && \multicolumn{2}{c}{Datasets}\\
        \cline{3-4}
        && FG-NET & CACD-VS \\
        \midrule  
        BaseFace && 90.55 & 98.48 \\
        DebFace && 93.3 & 99.45 \\
        \bottomrule
    \end{tabular}}
\end{table}

We also conduct experiments on two cross-age face recognition datasets, i.e., FG-NET~\footnote{\url{https://yanweifu.github.io/FG_NET_data}} and CACD-VS~\cite{chen2014cross}, to evaluate the age-invariant identity features learned by DebFace. The CACD-VS consists of 4,000 genuine pairs and 4,000 imposter pairs for cross-age face verification. On FG-NET, the evaluation protocol is the leave-one-out cross-age face identification. Table~\ref{tab:crossage} reports the performance of BaseFace and DebFace on these two datasets. Compared to BaseFace, the proposed DebFace improves both the verification accuracy on CACD-VS and the rank-1 identification accuracy on FG-NET.

\clearpage
%
%
{\small
\bibliographystyle{splncs04}
\bibliography{egbib}
}